\documentclass{spie}

\usepackage{graphicx}
\usepackage[cmex10]{amsmath}
\usepackage{amsfonts}
\usepackage{amssymb}
\usepackage{stmaryrd}
\usepackage{subfigure}
\usepackage{algorithm,algorithmic}
\usepackage{multirow}
\usepackage{array}
\usepackage{caption}

\def\R{{\mathbb{R}}}

\title{Detection and tracking of gas plumes in LWIR hyperspectral video sequence data}

\author{
Torin Gerhart$^\star$,
Justin Sunu$^\star$,
Lauren Lieu$^\ast$,
Ekaterina Merkurjev$^\dagger$,
Jen-Mei Chang$^\star$,
J\'{e}r\^{o}me Gilles$^\dagger$,
Andrea L. Bertozzi$^\dagger$,
}

\authorinfo{
\\$^\star$ Department of Mathematics and Statistics, California State University Long Beach, Long Beach, CA, USA
\\$^\dagger$ Department of Mathematics, University of California Los Angeles, 520 Portola Plaza, Los Angeles, CA, USA
\\$^\ast$ Department of Engineering, Harvey Mudd College, Claremont, CA. USA
}

\begin{document}
\maketitle

\begin{abstract}
Automated detection of chemical plumes presents a segmentation challenge. The segmentation problem for gas plumes is difficult due to the diffusive nature of the cloud. The advantage of considering hyperspectral images in the gas plume detection problem over the conventional RGB imagery is the presence of non-visual data, allowing for a richer representation of information. In this paper we present an effective method of visualizing hyperspectral video sequences containing chemical plumes and investigate the effectiveness of segmentation techniques on these post-processed videos. Our approach uses a combination of dimension reduction and histogram equalization to prepare the hyperspectral videos for segmentation. First, Principal Components Analysis (PCA) is used to reduce the dimension of the entire video sequence. This is done by projecting each pixel onto the first few Principal Components resulting in a type of spectral filter. Next, a Midway method for histogram equalization is used. These methods redistribute the intensity values in order to reduce flicker between frames. This properly prepares these high-dimensional video sequences for more traditional segmentation techniques. We compare the ability of various clustering techniques to properly segment the chemical plume. These include K-means, spectral clustering, and the Ginzburg-Landau functional.
\end{abstract}


\section{ Introduction }\label{sec:IT}
The detection of chemical plumes in the atmosphere is a problem that has significant applications to defense, security, and environmental protection. The accurate identification and tracking of airborne toxins is crucial to combat the use of chemical gases as weapons. The problem lies in the detection of these toxins. Standard images with red, green, and blue (RGB) intensity values are not able to capture the potentially invisible gas plume and thus an alternative data representation is required. Hyperspectral imaging is utilized for its ability to capture data in the non-visible electromagnetic spectrum, instead of three channels per pixel there could be over a hundred. These extra channels allow for the possible segmentation and detection of gas plumes; however it's offset by the fact that the images become computationally expensive to work with. In our work, we demonstrate a method to reduce the dimensions of a hyperspectral image, visualize a hyperspectral video sequence, and then perform image segmentation.

The first step to visualize the data is a reduction in dimension.  Within these hundreds of channels in a hyperspectral image, information of negligible importance can be found.  In order to remove these unimportant information, we perform Principal Component Analysis (PCA) in order to reduce the dimensions and retain most information.  Afterwards, we produce a false color movie representation of the data by selecting the three channels with most information content and setting those to be the intensity values of an RGB image. However, even slight variability of the channel signals between frames results as flicker in the video sequence. We utilize a Midway histogram equalization method to correct for this effect. This allows for a consistent pixel signature across the sequence of images. The result is a very smooth video sequence that allows for image classification and segmentation techniques.  The standard for chemical plume detection in hyperspectral imaging is to utilize target signatures and match areas that are closest to the target, such as automated matched subspace detection. While these techniques are able to isolate areas of the target signature, there is a high rate of false positives \cite{}. An alternate approach is image segmentation which allows for the inclusion of spectral correlation important to the detection of gas plumes. We use three different segmentation techniques: K-means, spectral clustering, and a modified MBO (Merrimen, Bence, and Osher) scheme that minimizes the Ginzburg-Landau functional. K-means is a well known clustering technique and can give useful information about the data. Spectral clustering is a technique that utilizes the graph Laplacian in order to find different clusters within the data. The last technique, a modified MBO scheme, is a semi-supervised segmentation method involving a diffusion process on a graph.

In section \ref{sec:BG} we will go over previous work that has been done in the detection of gas plumes in hyperspectral images.  From there, section \ref{sec:DP} goes into the data processing that is done.  Section \ref{sec:C} will be an overview of the clustering methods that are used.  Section \ref{sec:EXP} goes over the experimental results.

\section{ Background and Previous Work }\label{sec:BG}
\begin{figure}[h!]
	\centering
	\includegraphics[width=0.5\textwidth]{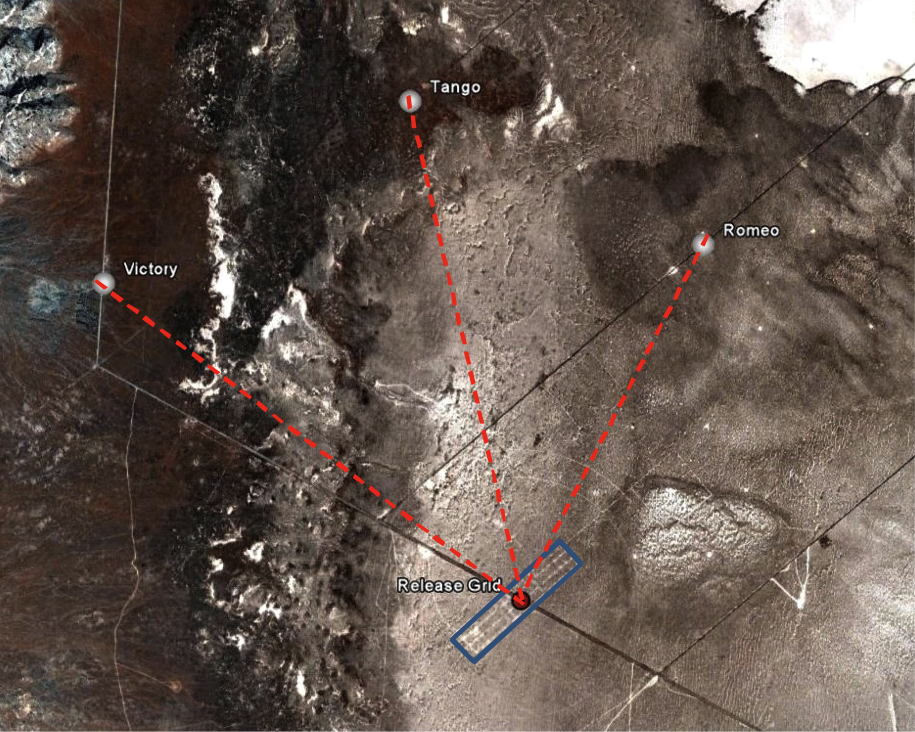}
	\caption{The locations of the Romeo, Victory, and Tango cameras, and the location of the chemical plume release.}
	\label{fig:RTV}
\end{figure}
\noindent The hyperspectral data set analyzed for this project was provided by the Applied Physics Laboratory at Johns Hopkins University. It consists of a series of video sequences recording the release of chemical plumes taken at the Dugway Proving Ground. Videos were captured by three long wave infrared spectrometers (named Romeo, Victory and Tango) placed at different locations to track the release of known chemicals. Each camera was about 2 kilometers away from each release at an elevation of about 1300 feet. Figure \ref{fig:RTV} shows the arrangement of the three cameras. The sensors capture one data cube every five seconds consisting of measurements at wavelengths in the long wave infrared (LWIR) portion of the electromagnetic spectrum. Each layer in the spectral dimension depicts a particular frequency starting at 7,830 nm and ending with 11,700 nm with a channel spacing of 30 nm. The spatial dimension of each of these data cubes is 128 $\times$ 320 pixels, while spectral dimension is 129 channels.

\subsection{Data Conversion} \label{sec:DC}
The first step in dealing with hyperspectral data is to convert from spectral radiance to emissivity. This is done because most gaseous plumes have little to no spectral radiance. The emissivity of a surface is a measurement of the light leaving a surface relative to that of a perfect blackbody, ranging between 0 and 1. Planck\textquoteright s blackbody equation directly relates the spectral response of a surface to emissivity and temperature \cite{Carr2011},
\begin{equation}  
	B(\nu,T)=\frac{2hc^2\nu^3 }{\exp(\frac{hc\nu}{kT})-1}
\end{equation}
where $B(\nu,T)$ is the spectral excitance at a given wavenumber $\nu$ and temperature $T$, $h$ is Planck\textquoteright s constant, $c$ is the speed of light, and $k$ is Boltzmann\textquoteright s constant. This calculation assumes that the entire scene is a constant temperature, which is inaccurate. There are at least three dominant temperatures present in the image: the atmosphere, the distant mountains, and the desert foreground. Due to this irregularity, outliers are formed in the emissivity data, outside of the expected 0 to 1 emissivity range. In order to clean the data, a 3 $\times$ 3 spectral median filter is implemented on the outlier pixels prior to running the plume detection and segmentation algorithms.

\subsection{ Adaptive Matched Subspace Detector } \label{sec:AMSD}
\begin{figure}[h!]
	\centering
	\includegraphics[width=0.5\textwidth]{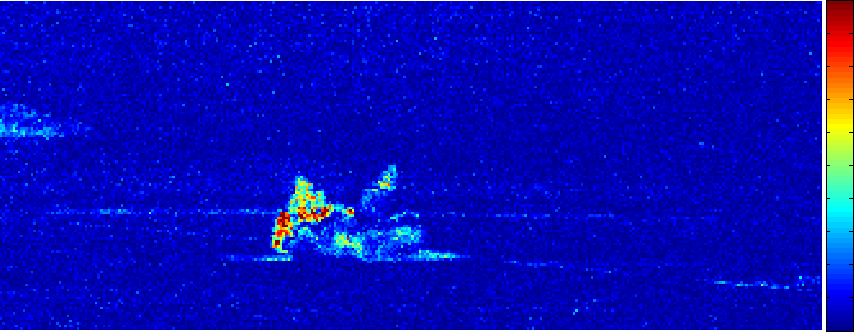}
	\caption{A resulting frame after applying the AMSD.  Red indicates higher probability of being target signature.  Note the numerous areas outside of the gas plume that could be miscatergorized as gas.}
	\label{fig:amsd}
\end{figure}

The Adaptive Matched Subspace Detector\cite{AMSD} (AMSD) is a probabilistic detection scheme that uses a generalized likelihood ratio test to choose between the hypotheses \\ ~ \\
\begin{equation}
	\label{eq:AMSD}
	\begin{aligned}
		& H_0 : x = S_b u_b + n & \text{(Target absent)} \\
		& H_1 : x = S_t u_t + S_b u_b + n = S x + n & \text{(Target present)} \\
	\end{aligned}
\end{equation}
\noindent where $n \sim \mathcal{N}(0,\sigma_w^2 I)$, $\mathcal{N}$ is the Gaussian distribution, $x$ is a pixel in the image, $S_t$ is a matrix of target signatures, $S_b$ is a matrix of background signatures, and $u$ is a vector of abundances of the spectral signatures in $S$. The first hypothesis, $H_0$, corresponds to the situation when the pixel $x$ may be represented by only background pixels plus noise. The hypothesis $H_1$ says that the target signature is needed in addition to the background to fully represent the pixel. A hypothesis test is constructed using a generalized likelihood ratio, given by
\begin{equation}
	\mathcal{L}(x) = \bigg(\frac{x^T P^{\perp}_b x }{x^T P_S^{\perp} x} \bigg)^{L/2}
\end{equation}
where the matrix $P_S$ is the projection onto the subspace spanned by S, $L$ is the number of spectral components, and $P_b$ is the projection onto the subspace spanned by the background.  The value of $\mathcal{L}(x)$ determines the likelihood of a pixel $x$ containing the target signature. In order to determine if a pixel contains the target, $\mathcal{L}(x)$ is compared to a given threshold $l_0$. If $\mathcal{L}(x) < l_0$, then pick $H_0$, otherwise pick $H_1$. In order to ensure linear independence of the numerator and denominator (that is, they are uncorrelated) the ratio \\
\begin{equation}\label{eq:prob}
	T_{\textnormal{AMSD}}(x) = \frac{x^T ( P_b^{\perp} - P_S^{\perp}) x}{x^T P_S^{\perp} x} = \mathcal{L}(x)^{L/2} - 1
\end{equation}
 is used. The resulting probability distribution of the AMSD on a frame can be seen in Figure \ref{fig:amsd}.  We will be using the AMSD as a benchmark to test our clustering results.  Further information on the probability distribution associated with equation \eqref{eq:prob} may be found in \cite{AMSD}.

\section{ Data Processing } \label{sec:DP}
After the data conversion into spectral emissivity, the signatures of the gas should become prevalent.  However, the data is still full dimension, meaning of dimension 128 $\times$ 320 $\times$ 129, and clustering algorithms become computational expensive with large datasets.  We also need to take into account how the pixel values vary due to temperature fluctuations during the day.  If we choose to utilize a video segmentation technique, there needs to be a smooth transition from frame to frame.  All of these are taken into account in our data processing; dimension reduction to reduce the data and Midway equalization to smooth frame transitions.

\subsection{ Dimension Reduction } \label{sec: }
\begin{figure}[h]
\centering
\includegraphics[width=0.48\textwidth]{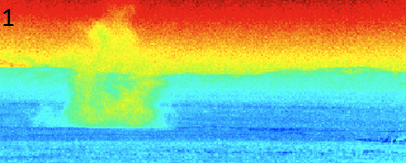}
\includegraphics[width=0.48\textwidth]{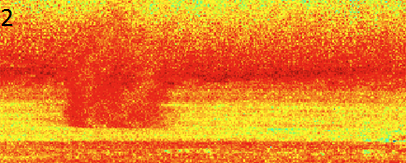}
\vskip 0.01\textheight
\includegraphics[width=0.48\textwidth]{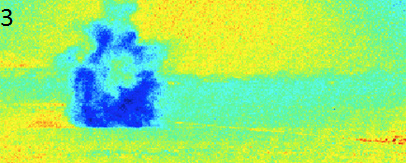}
\includegraphics[width=0.48\textwidth]{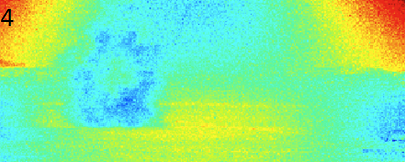}
\vskip 0.01\textheight
\includegraphics[width=0.48\textwidth]{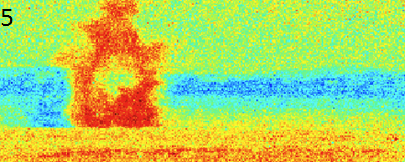}
\caption{The first five principal components resulting from PCA.  The first image is the first principal component, second image is second component, etc. Note how the first, third, and fifth components provide the most contrast between the background and gas plume.}
\label{fig:pca}
\end{figure}

Segmentation techniques require the creation of rather large similarity matrices. The computational cost of these algorithms grows as the dimension of the data increases. There are numerous methods to reduce the dimension of data; we choose to use Principle Components Analysis (PCA) in this work since it is widely known and well understood. Briefly, a data matrix, $X$, of the video sequence is constructed by treating each pixel as a vector in $\R^{129}$ and arranging all the pixels into columns. Then the principle components are found as eigenvectors of the covariance matrix $XX^T$. By performing PCA in this way, the principle components act as a type of spectral filter where each component represents a particular chemical spectrum in the image \cite{Mar2006}. This is sometimes referred to as \textit{endmember extraction}. In order to obtain a $k$-dimensional representation of the data each pixel is projected onto $k$ principle components. False color RGB videos were created by treating three chosen projections as a 3-dimensional representation of the data, forming a red, green, and blue intensity values. In our video, we chose to utilize the first, third, and fifth projected component because these provided the best contrast between plume and background.  The first five projections, are shown in Figure \ref{fig:pca}.  However, this false color video has large fluctuations in intensity values between frames. These fluctuations appear as flickering in the video sequence and may be seen in the left column of Figure \ref{fig:midway}.

\subsection{ Midway Equalization } \label{sec:ME}
The individual false color RGB frames generated by PCA are relatively free of noise but there is flickering inconsistency throughout the RGB video as a whole. The values for each color component shift from frame to frame.  This may be attributed to noise in the original data, variance in the data from frame to frame, and possible artifacts from the LWIR sensor. To correct for these fluctuations in intensity, the Midway equalization method\cite{Delon2004} was utilized. The Midway equalization method is used to equalize the histograms of each frame in the video sequence. This is done by averaging the histograms of every frame in the video sequence, then replacing the histogram of each frame by the average. This causes the frames to have the same distribution of light and dark values, which eliminates any flicker between frames. As a byproduct of redistributing light values in each frame, contrast between certain light and dark areas is reduced. This adverse effect may be reduced by using contrast enhancement methods before or after Midway equalization.

\subsection{ Results } \label{sec:mwresults }

\begin{figure}[h]
\centering
	\begin{tabular}{c|c}
    PCA False Color Video Frames & Midway Equalized Video Frames \\
		\includegraphics[width = .48\textwidth]{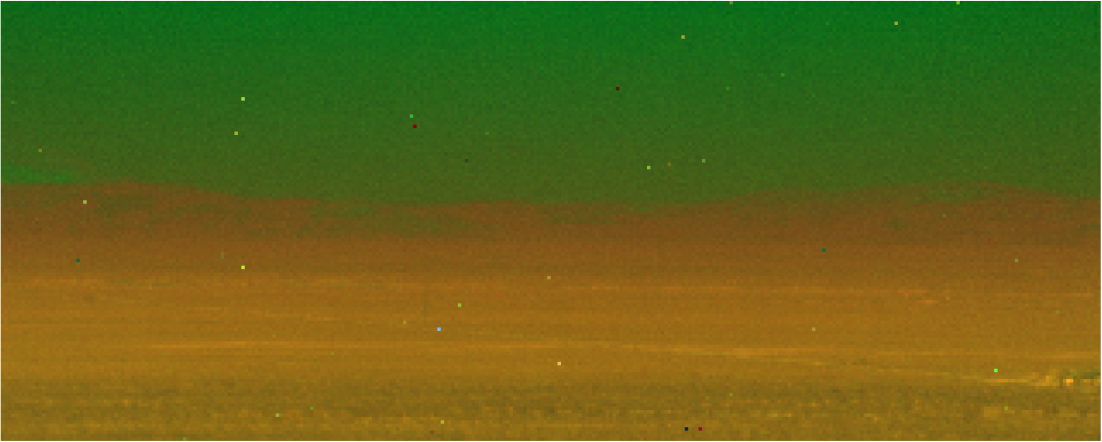} &
		\includegraphics[width=.48\textwidth]{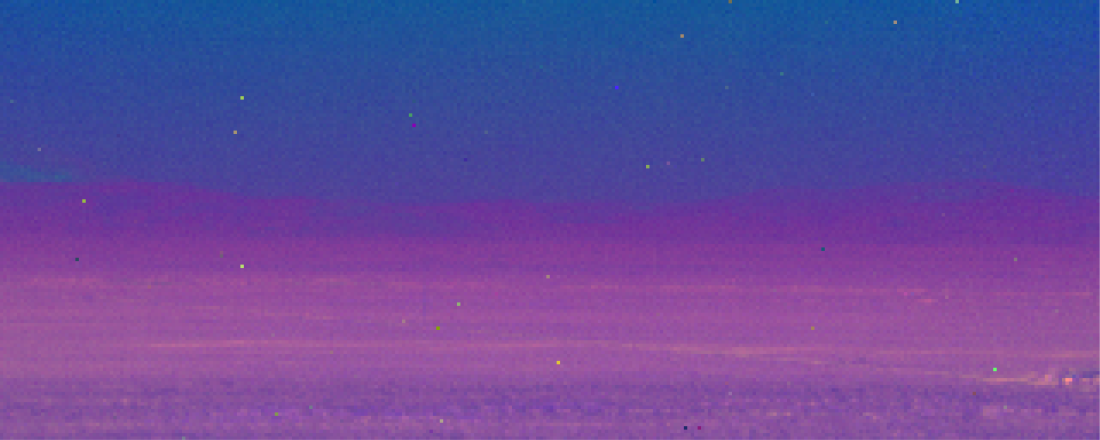} \\
		\includegraphics[width=.48\textwidth]{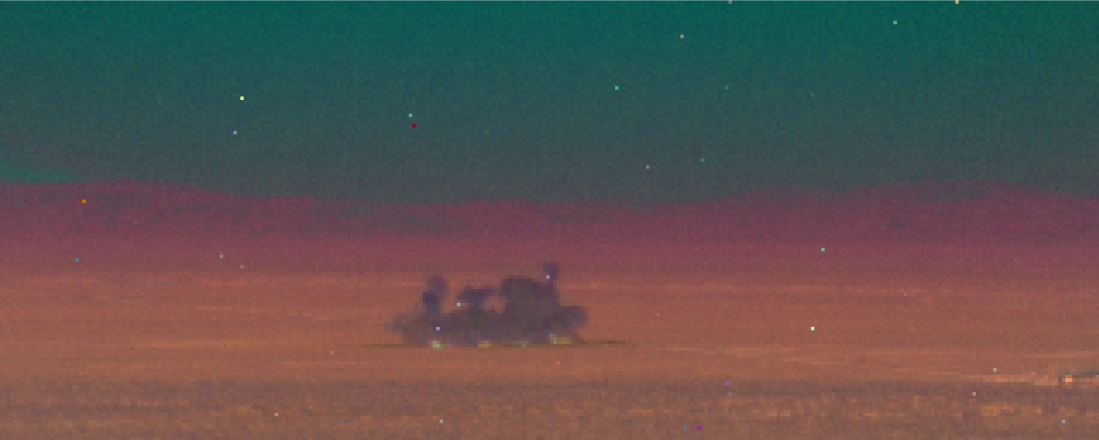} &
		\includegraphics[width=.48\textwidth]{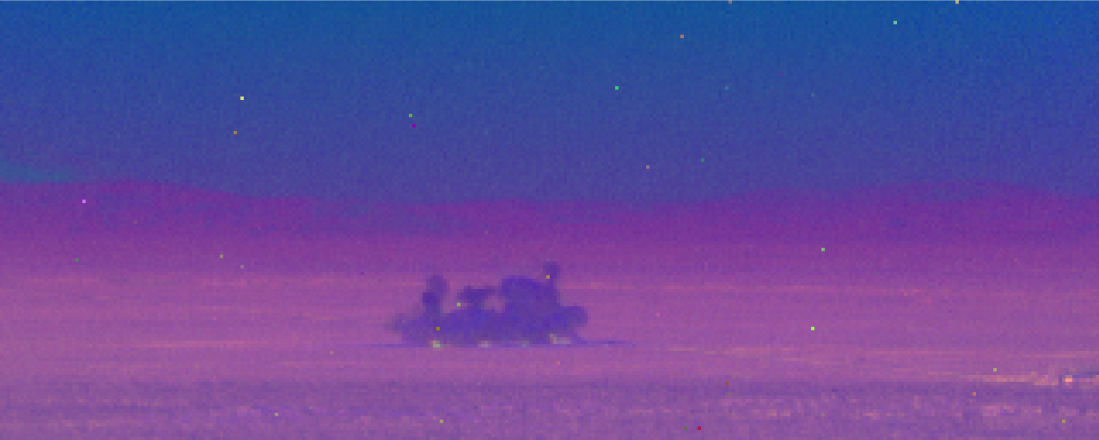} \\
		\includegraphics[width=.48\textwidth]{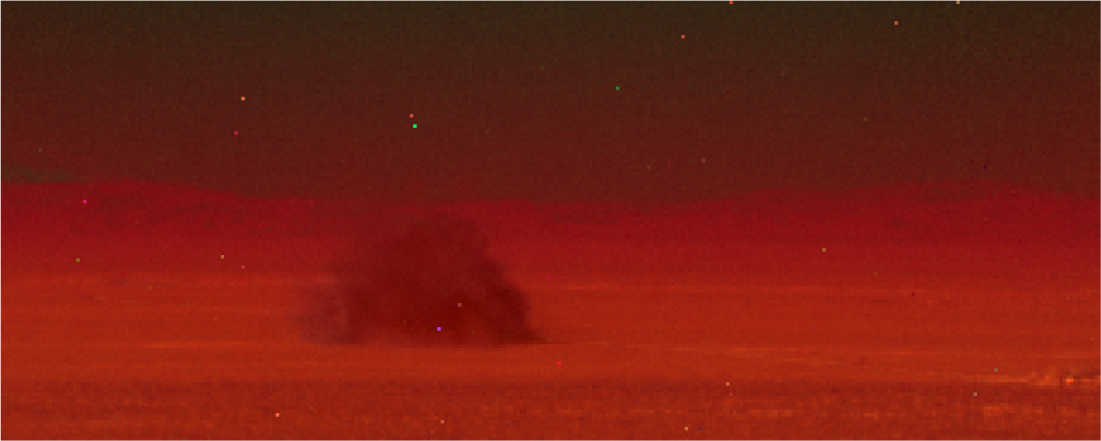} &
		\includegraphics[width=.48\textwidth]{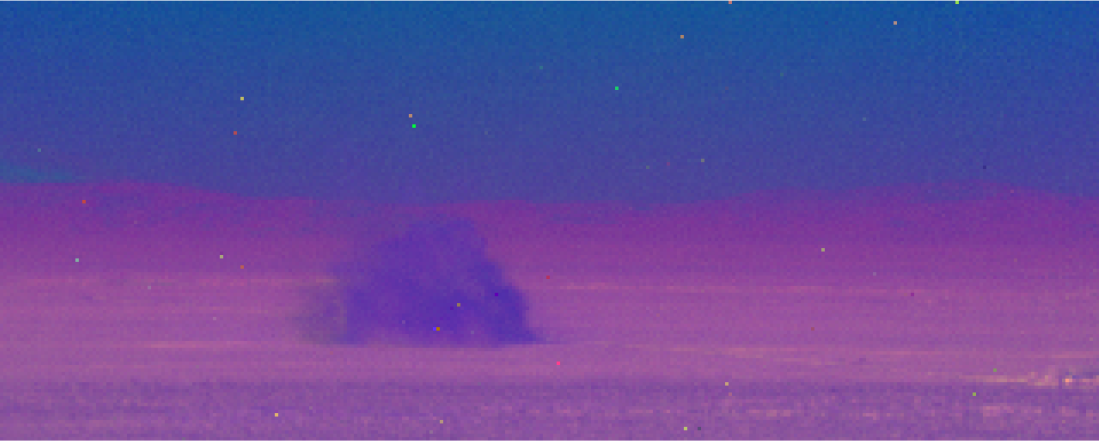} \\
	\end{tabular} \caption{Results of the Midway equalization on the false color video sequence.  Colors are created using false color mapped from the first, second, and third principal components.  Each frame is at a different time step, illustrating the color fluctuations and gas plume release.  The left column has the images with the color variance.  The right column has the images resulting from applying the Midway equalization algorithm.}
	\label{fig:midway}
\end{figure}
\begin{figure}
  \centering
  \includegraphics[width=.48\textwidth]{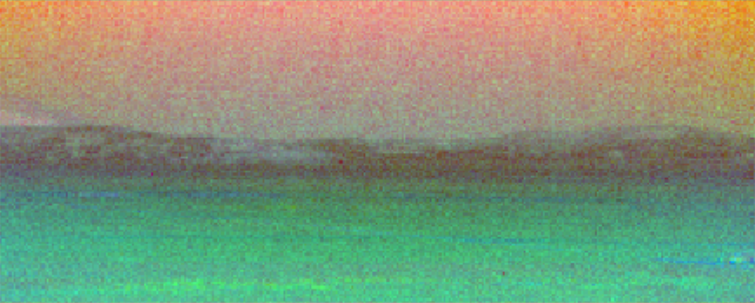} \\
  \includegraphics[width=.48\textwidth]{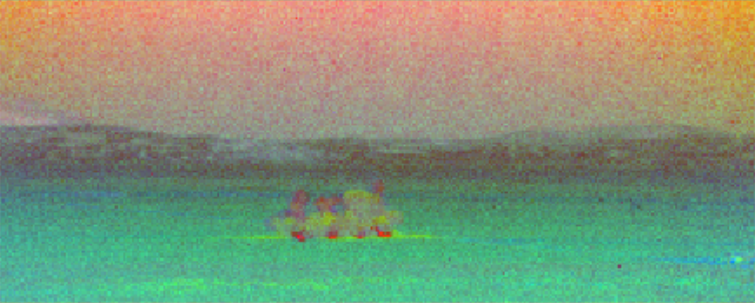} \\
  \includegraphics[width=.48\textwidth]{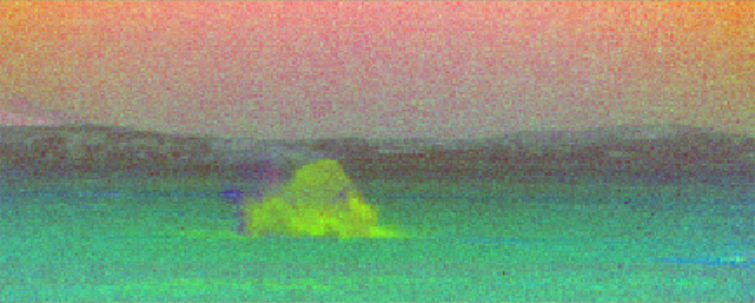}
  \caption{Midway equalized false color video sequence utilizing first, third, and fifth principal components.  This is the video with the principal components that provided the most contrast between gas plume and background. }
  \label{fig:monet}
\end{figure}

\noindent Figure \ref{fig:midway} shows the resulting false color frames after the application of Midway equalization. The left side of the figure shows the original false color video, and the right side shows the results of Midway Equalization. The original video frames each has a different coloring, varying from yellow and green to completely red. After the equalization, each of the three images has similar shades of purple and pink. This provides continuity for the video sequence and displays movement of the plume more clearly as the scene changes from frame to frame. For the purposes of our clustering, we chose to utilize a video sequence with the first, third, and fifth principal components.  These provided the highest contrast between the foreground and gas plume.  The resulting false color video frames are shown in Figure \ref{fig:monet}.

\section{ Clustering } \label{sec:C}
As mentioned in section \ref{sec:AMSD}, AMSD produces a probability distribution for the presence of a target spectral signature. The goal is to isolate the plume without a target spectral signature, because this may not always been known in applications. With this reduction and smoothing of the data, more conventional clustering methods can be utilized, in this case spectral clustering and a modified MBO scheme.  In the following sections we will go over the clustering methods that were used and in section \ref{sec:EXP} the results will be covered.

Two different distance metrics will be utilized for the clustering, cosine (spectral angle) and Euclidean distance. For $\textbf{x}$ and $\textbf{y}$ representing a pixel's spectral signature, these metrics are defined as:
\begin{equation}
	\begin{aligned}
		d_{c} (\textbf{x},\textbf{y})& = 1-\frac{<\textbf{x},\textbf{y}>}{||\textbf{x}||~||\textbf{y}||} & \text{(cosine)} ,
	\end{aligned}
	\label{eq:cos}
\end{equation}

\begin{equation}
	\begin{aligned}
		d_{e}(\textbf{x},\textbf{y}) & = \sqrt{<\textbf{x}-\textbf{y},\textbf{x}-\textbf{y}>} & \text{(Euclidean)} .
	\end{aligned}
	\label{eq:euc}
\end{equation}

\begin{figure}[h!]
\centering
	\includegraphics[width = .48\textwidth]{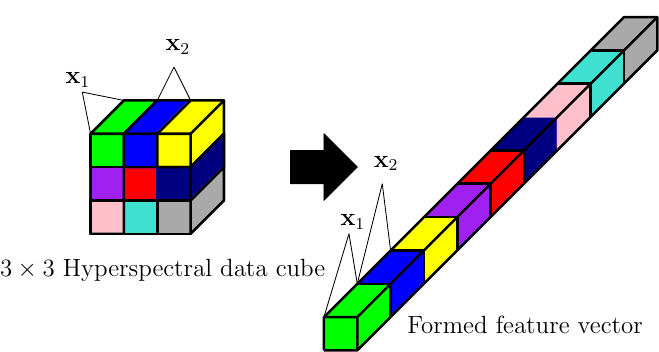}
	\caption{An illustration of how a feature vector is formed. The formed feature vector is denoted by $\mathbf{x}$ in the text.}
	\label{fig:feve}
\end{figure}

These clustering methods utilize information present at a pixel location and try to form groups with similar pixels.  However, this approach fails to utilize information that is present in the surrounding pixels. In order to incorporate this spatial information, we utilize feature vectors. To form a feature vector, replace a pixel with the concatenated $3 \times 3$ surrounding pixels. This allows for a comparison of the surrounding information of each pixel.  A demonstration of this process is shown in Figure \ref{fig:feve}.  In this figure, $\mathbf{x}_i$ represents a pixel's spectral signature.

With the incorporation of feature vectors, we need to modify the cosine metric. The reason for forming feature vectors is to compare corresponding spatial information; however, the cosine metric would just calculate the spectral angle between the two formed vectors. The modified cosine metric uses the spectral angle for the spectral data at each pixel and uses the 2-norm across spatial locations:

\begin{equation}
	\begin{aligned}
d_{\overline{c}}(\mathbf{x},\mathbf{y}) = \sqrt{\sum^9_{i=1} \left( 1 - \frac{<\textbf{x}_i,\textbf{y}_i>}{||\textbf{x}_i||~||\textbf{y}_i||} \right) ^2}
 = \sqrt{\sum^9_{i=1} \left( d_c (\textbf{x}_i,\textbf{y}_i) \right) ^2} & \text{ (modified cosine)} .
 	\end{aligned}
	\label{eq:mcos}
\end{equation}

\subsection{ K-means} \label{sec:KM}
K-means is the first clustering technique that we performed on the data. The K-means algorithm is summarized in Algorithm \ref{algo:KMAlgo}. While it does not provide definitive clustering results, it is useful for deriving information about the data set at very little cost. Testing is done on the false color frames as well as the feature vector frames. Specifically we test the different distance metrics, viability of feature vectors, and presence spatial information. There are two central problems with K-means. The first problem is how the initial selection of centroids can greatly change the outcome of clustering.  This means having two different initial centroids can result in two completely different resulting segmentations. The second is having a $k$ value different from the actual number of clusters in the data. The first problem can result in awkward clusterings while the second problem will either form more clusters or fuse clusters together.

\begin{algorithm}
	\caption{K-means Algorithm}
	\label{algo:KMAlgo}
	\begin{algorithmic}
		\STATE $k$ is the number of clusters, $I$ is image data with $n$ spectral components, $I_{ij}$ is a pixel in the image, $I_{ij} = (I_{(ij,1)},I_{(ij,2)},...,I_{(ij,n)})$
		\STATE If not provided, randomly select pixels as initial centroids
		\STATE Calculate distance from each pixel to each centroid
		\STATE Assign each pixel to the closest centroid
		\STATE If: Each pixel remains unchanged with its pixel assignment, then K-means is finished
		\STATE Else: Recalculate centroids and start over
	\end{algorithmic}
\end{algorithm}

\subsection{ Spectral Clustering } \label{sec:SC}

Due to the problems in K-means, an alternative clustering algorithm is required, one that does not rely so heavily on initialization parameters. Spectral clustering is an unsupervised technique that utilizes the eigenvectors of the Laplacian matrix to form clusters. There are a few variations of the spectral clustering algorithm that differ by what eigenvalues and eigenvectors are used. The method of Von Luxburg \cite{Luxburg2007} utilizes the eigenvectors corresponding to the smallest eigenvalues to form clusters, whereas the method of Jordan and Weiss\cite{Ng2002} uses the largest corresponding eigenvectors. Since the computation of the the smallest corresponding eigenvectors tends to be more time consuming than computing the largest corresponding eigenvectors, we chose to use the method of Jordan and Weiss. This spectral clustering algorithm is summarized in Algorithm \ref{algo:SCAlgo}. Spatial information is also tested, utilizing feature vectors, so the modified cosine and standard Euclidean are used as the distance metrics.
\begin{algorithm}
	\caption{Spectral Clustering}
	\label{algo:SCAlgo}
	\begin{algorithmic}
		\STATE $I$ is image data with $n$ spectral components, $I_{ij}$ is a pixel in the image, $I_{ij} = (I_{(ij,1)},I_{(ij,2)},...,I_{(ij,n)})$
		\STATE Form $ D $, the matrix of distances between each pixel
		\STATE Form $ S $, the matrix of similarities between each pixel
		\STATE Calculate $d$, the matrix of row sums. $d_{ii} = \sum^n_{j=1}S_{ij}$
		\STATE Form $ N $, the normalize similarity matrix, $d^{-0.5} S d^{-0.5}$
		\STATE Compute the largest eigenvalues and corresponding eigenvectors of matrix $N$
	\end{algorithmic}
\end{algorithm}

The similarity between two points is inversely proportional to the distance, i.e., the smaller the distance the higher the similarity and the larger the distance the smaller the similarity.  The range of similarity is between zero and one, where a value of zero means no similarity and a value of one gives an identical match.  The standard similarity function is the Gaussian similarity, which utilizes the computed distances and converts them into a Gaussian distribution. Therefore, the similarity matrix is given by

\begin{equation}\label{eqn:Sim}
	S_{ij} = e^{-\frac{d(x_i, x_j)^2}{\sigma ^2}}
\end{equation}

\noindent where $\sigma \in \R^{+}$ is a chosen parameter and $d(x_i,x_j)$ is one of the aforementioned distance metrics. After experimentation, we found that $\sigma = 1$ produces decent results for both distance metrics on this particular data set.

After normalization, compute the largest eigenvalues and corresponding eigenvectors.  The first, or the largest, eigenvalue and corresponding eigenvector provides no clustering results, but gives the mean of the data.  So aside from the first, these eigenvalues describe the relevance of the resulting clusters, and the corresponding eigenvectors describe the clusters.

\subsection{ Modified MBO Scheme for Segmentation } \label{sec:MBO}

Spectral clustering provides a better view of the different aspects to the hyperspectral images, however it is not able to exclusively isolate the gas plume because it is an unsupervised method.  The MBO scheme is a semi-supervised clustering method that will allow for the segmentation of the gas plume. The Ginzburg-Landau functional is defined as
\begin{equation} \label{eq:GL}
	GL(u) = \frac{\epsilon}{2}\int{|\nabla u|^2 dx} + \frac{1}{\epsilon}\int{W(u)dx}
\end{equation}
where $W(u)$ is a double well potential, such as $W(u) = (u^2 - 1)^2$. When minimizing Equation \eqref{eq:GL} the double well potential term will force solutions to the minimizers of $W(u)$. The first term in Equation \eqref{eq:GL}, containing the spatial gradient operator $\nabla$, will incorporate smoothness into the solution. Therefore, any sharp transition between the two minimizers of $W(u)$ will be smoothed out. The solution that minimizes \eqref{eq:GL} will have regions close to one of the minimizers of $W(u)$, as well as an interface between the two regions. Models with this property are referred to as ``diffuse interface" models. \\

In order to use this technique to segment an image, this method is modified to work on a graph. To do this the gradient term is replaced with a non-local, graph version of the discretized Laplace operator, $\epsilon u \cdot L_s u$. The normalization of the graph Laplacian used here is given by, \\
\begin{equation} \label{eq:GLap}
L_s = I - d^{-1/2} W d^{-1/2}
\end{equation}
and is referred to as the \textit{symmetric Laplacian}, as the result is a symmetric matrix.  The matrix $d$ is the same as defined in Algorithm \ref{algo:SCAlgo}. Other normalizations of the graph Laplace operator are discussed in \cite{BF2012}. After constructing $L_s$, a fidelity term is added to the functional, and is given by,
\begin{equation} \label{eq:BF}
GL(u) = \epsilon u \cdot L_s u + \frac{1}{\epsilon}\int{W(u) \, dx} + \int{F(u, u_0)}
\end{equation}
where $F(u, u_0)$ is the additional fidelity term. This is a semi-supervised segmentation algorithm and is initialized with a patch of the region of interest, $u_0$. As the diffusion process is evolved, pixels similar to the initialized region will be clustered together. The paper by Merkurjev et al. proposes a modification of the MBO scheme in order to minimize the Ginzburg-Landau functional on a graph \cite{MBO2012}. The method is a two-step process, where the first step is solving the heat equation on a graph, and the second step is thresholding. More specifically, \\
\begin{enumerate}
\item Compute $y(x) = S(\delta t) u_n(x)$, where $S(\delta t)$ is the evolution operator of the equation \\
	\begin{equation*}
		\frac{\partial u}{\partial t} = -L_s u - C_1 \lambda (x) (u - u_0),
	\end{equation*}
\item Threshold, \\
	\begin{equation*}
		u_{n+1}(x) = \left\{
			\begin{array}{c c}
				 1 & \quad \text{if } \; u(x) \ge 0 \\
				-1 & \quad \text{if } \; u(x) < 0 . \\
			\end{array} \right. 
	\end{equation*}
\end{enumerate}
The equation in step one is essentially a diffusion equation with the additional term $- C_1 \lambda (x) (u - u_0)$ corresponding to the data-fidelity term added in Equation \eqref{eq:BF}. The parameter $\lambda$(x) determines how much of the initialized region is retained by the fidelity term. This is repeated for a prescribed number of iterations, or until $\frac{\| u_{n+1} - u_{n} \|_2^2}{\| u_{n+1} \|_2^2} < 10^{-6}$ is satisfied. A full discussion of this method may be found in \cite{MBO2012}.\\

\section{Experimental Results} \label{sec:EXP}
For our experiment,  we tried to utilize hyperspectral data with the goal of finding a gas plume release.  Romeo aa13 data set is considered in our experiments. The third frame after the plume was released is used for K-means and spectral clustering; while the first 40 frames are used for the modified MBO scheme. The work was done on a 2.4 Ghz core2duo processor with 4 GB of ram.  All implementation was done utilizing MATLAB.

We decided to utilize the false color videos that utilized the first, third, and fifth principal component.  We also used feature vectors formed from a 3 $\times$ 3 patch around a pixel. Table \ref{tab:EXP} provide a list of all the computational results presented and their computational run time.

\begin{table}[h]
\centering
\begin{tabular}{|c|c|c|c|} \hline
Method & Distance Metric & Results & Time \\ \hline
\multirow{4}{*}{K-means} & Euclidean & Figure \ref{fig:kmean} & $<$ 1 second \\
 & Cosine & Figure \ref{fig:kmean} & $<$ 1 second \\
 & Euclidean with feature vectors & Figure \ref{fig:kmean} & $\sim$ 5 seconds \\
 & Modified cosine with feature vectors & Figure \ref{fig:kmean} & $\sim$ 5 seconds \\ \hline
\multirow{3}{*}{Spectral Clustering} & Euclidean & Figure \ref{fig:SCEuc} & 2 $\sim$ 3 minutes \\
 & Cosine & Figure \ref{fig:SCCos} & 2 $\sim$ 3 minutes \\
 & Modified cosine with feature vectors & Figure \ref{fig:SCCosFV} & 3 $\sim$ 4 minutes \\ \hline
\multirow{3}{*}{Modified MBO} & Initialization & Figure \ref{fig:INI} & $<$ 1 second \\
& Cosine on first five principal component & Figure \ref{fig:MBO5} & $\sim$ 7 seconds \\
& Cosine on 1-3-5 false color video & Figure \ref{fig:MBO3} & $\sim$ 7 seconds \\ \hline
\end{tabular}
\caption{Table of different methods and distance metrics.}
\label{tab:EXP}
\end{table}

\subsection{K-means}
\begin{figure}[h!]
	\centering
	\includegraphics[width = .48\textwidth]{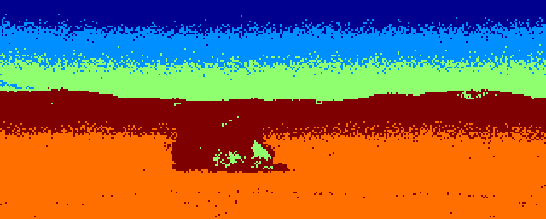}
	\includegraphics[width = .48\textwidth]{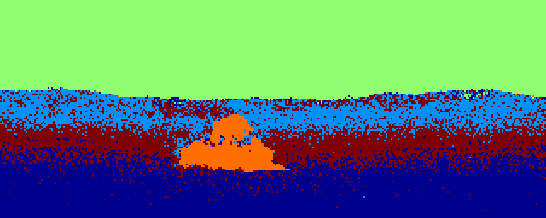}
	\vskip 0.01\textheight
	\includegraphics[width = .48\textwidth]{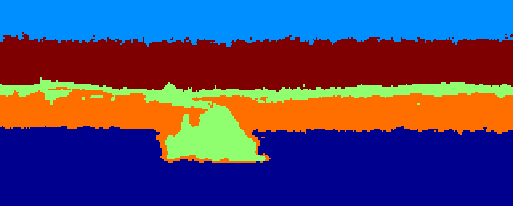}
	\includegraphics[width = .48\textwidth]{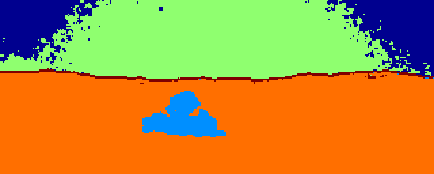}
	\caption{Results of K-means clustering with various distance metrics.  The cosine distance metric is able to isolate the plume in both cases; however, feature vectors give more defined boundaries.  Top left: Euclidean distance. Top right: Cosine metric. Bottom left: Euclidean with feature vectors. Bottom right: Modified cosine with feature vectors.}
	\label{fig:kmean}
\end{figure}

From the results, in Figure \ref{fig:kmean}, the cosine distance metric is able to isolate the plume quite clearly, although both metrics have a lot of bleeding between the different components in the image.  This is most likely due to how the hyperspectral signature ends up being a mixture of the two different areas.  With feature vectors, the resulting clusters provide a much cleaner differentiation between the different objects in the image.  Again the Euclidean distance is not able to isolate the plume, whereas the modified cosine results in a very distinct plume location.  From these results, we conclude that the cosine distance metric is preferred in isolating the plume.

\subsection{Spectral Clustering}
\begin{figure}
	\centering
	\includegraphics[width = .25\textwidth]{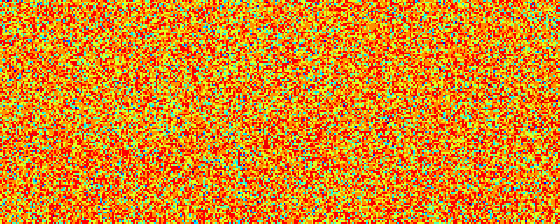}
	\includegraphics[width = .25\textwidth]{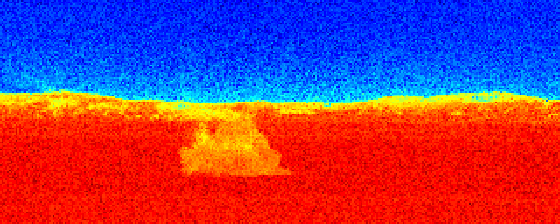}
	\includegraphics[width = .25\textwidth]{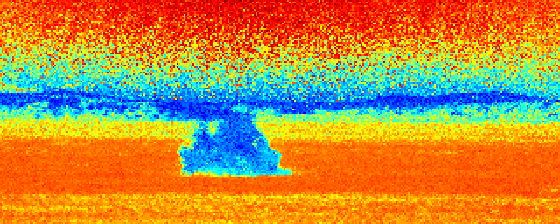}
	\vskip 0.005\textheight	
	\includegraphics[width = .25\textwidth]{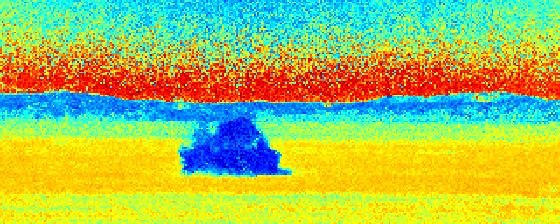}
	\includegraphics[width = .25\textwidth]{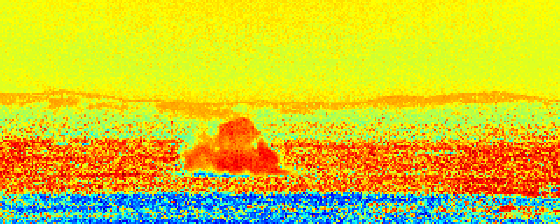}
	\includegraphics[width = .25\textwidth]{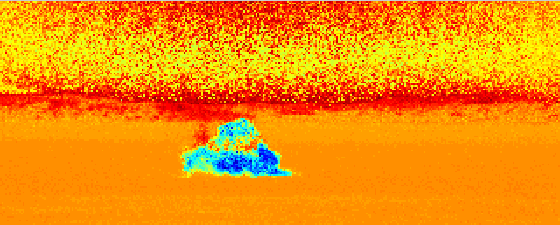}
	\vskip 0.005\textheight
	\includegraphics[width = .25\textwidth]{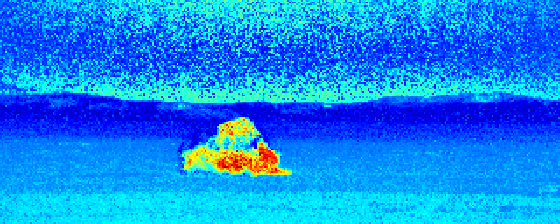}
	\includegraphics[width = .25\textwidth]{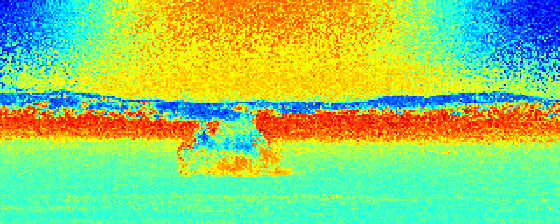}
	\includegraphics[width = .25\textwidth]{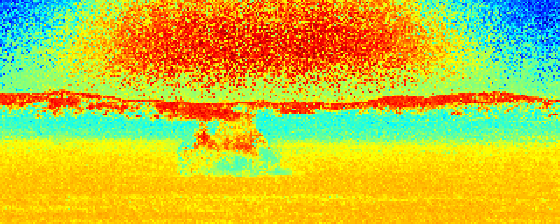}
	\vskip 0.005\textheight
	\includegraphics[width = .25\textwidth]{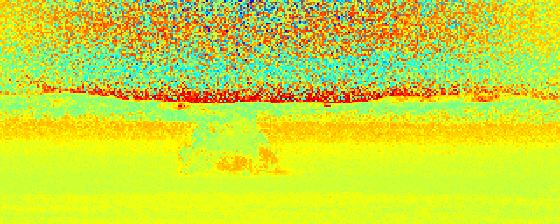}
	\includegraphics[width = .25\textwidth]{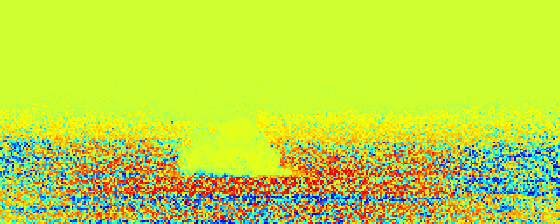}
	\includegraphics[width = .25\textwidth]{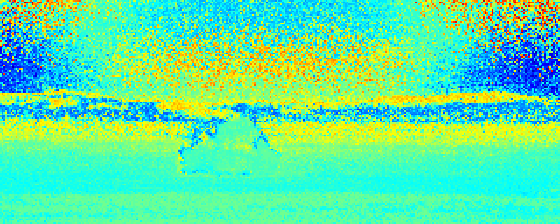}
	\vskip 0.005\textheight
	\includegraphics[width = .25\textwidth]{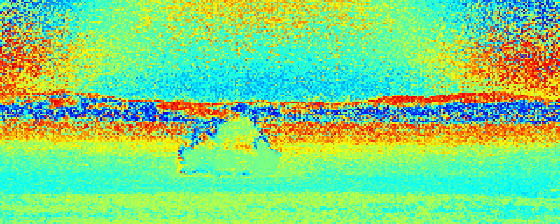}
	\includegraphics[width = .25\textwidth]{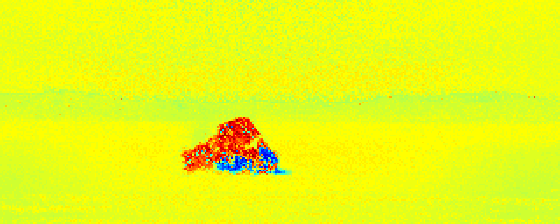}
	\includegraphics[width = .25\textwidth]{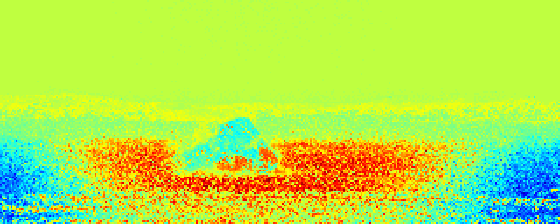}\\
\caption{Spectral clustering result: First 15 eigenvectors utilizing cosine distance metric.}
\label{fig:SCCos}
\end{figure}

\begin{figure}
  \centering
	\includegraphics[width = .25\textwidth]{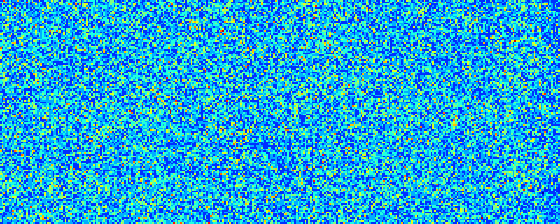}
	\includegraphics[width = .25\textwidth]{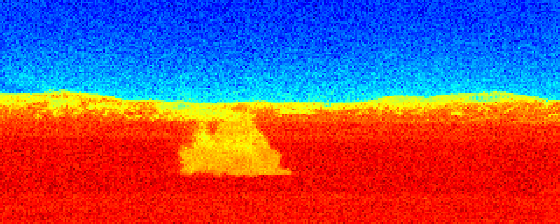}
	\includegraphics[width = .25\textwidth]{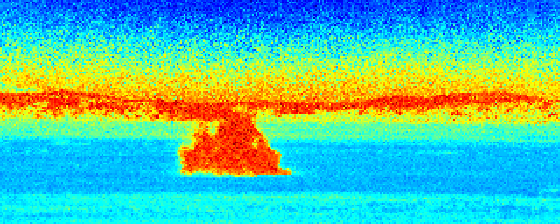}
	\vskip 0.005\textheight
	\includegraphics[width = .25\textwidth]{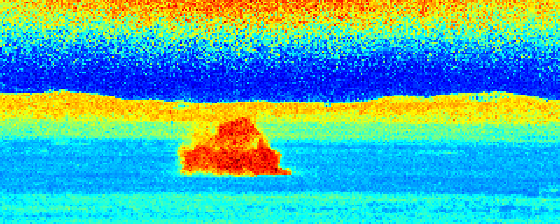}
	\includegraphics[width = .25\textwidth]{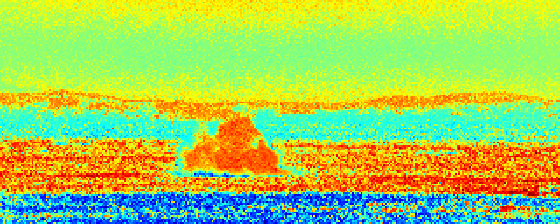}
	\includegraphics[width = .25\textwidth]{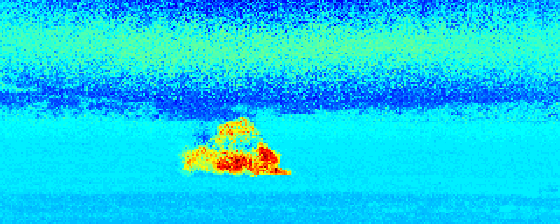}
	\vskip 0.005\textheight
	\includegraphics[width = .25\textwidth]{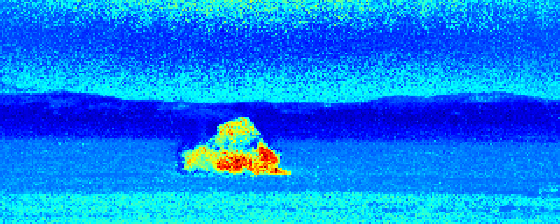}
	\includegraphics[width = .25\textwidth]{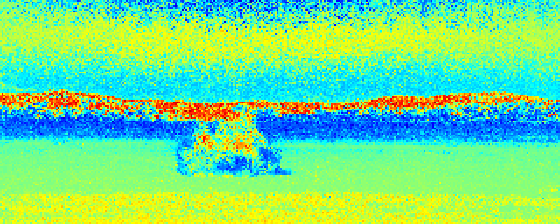}
	\includegraphics[width = .25\textwidth]{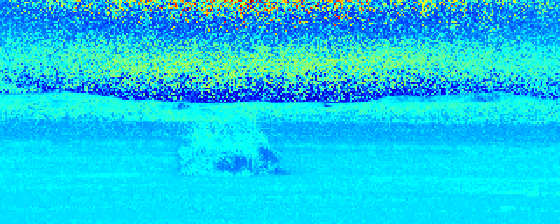}
	\vskip 0.005\textheight
	\includegraphics[width = .25\textwidth]{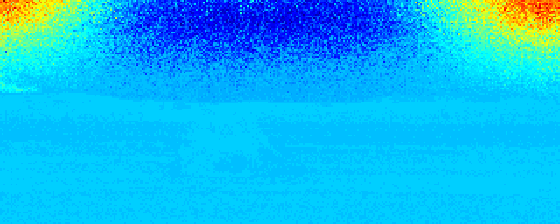}
	\includegraphics[width = .25\textwidth]{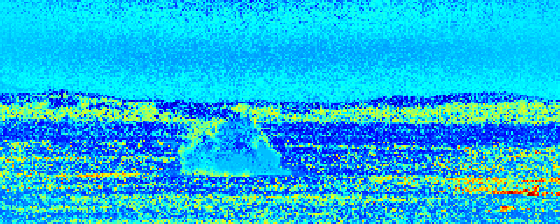}
	\includegraphics[width = .25\textwidth]{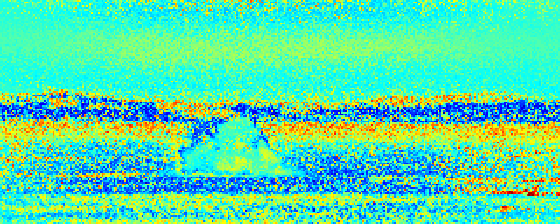}
	\vskip 0.005\textheight
	\includegraphics[width = .25\textwidth]{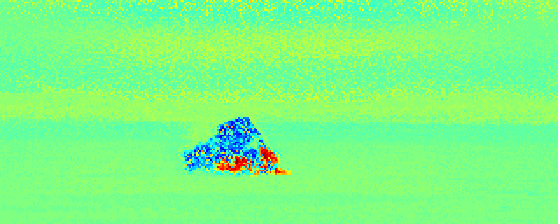}
	\includegraphics[width = .25\textwidth]{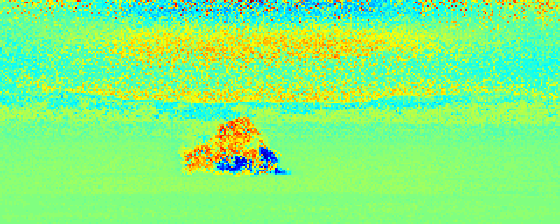}
	\includegraphics[width = .25\textwidth]{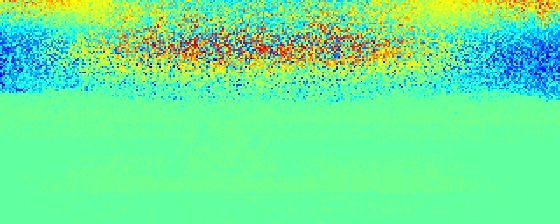}\\
	\caption{Spectral clustering result: First 15 eigenvectors utilizing Euclidean distance metric.}
	\label{fig:SCEuc}
\end{figure}

\begin{figure}
  \centering
	\includegraphics[width = .25\textwidth]{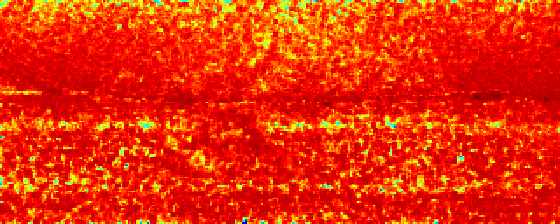}
	\includegraphics[width = .25\textwidth]{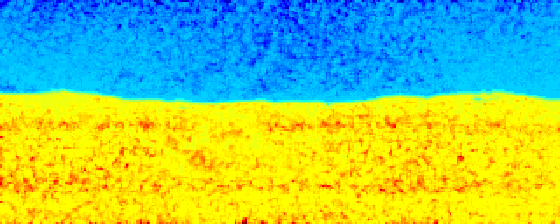}
	\includegraphics[width = .25\textwidth]{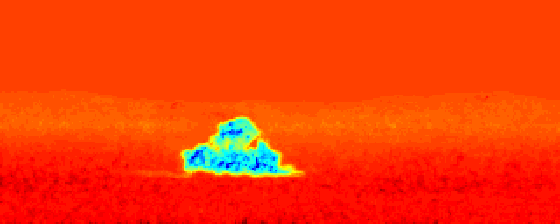}
	\vskip 0.005\textheight
	\includegraphics[width = .25\textwidth]{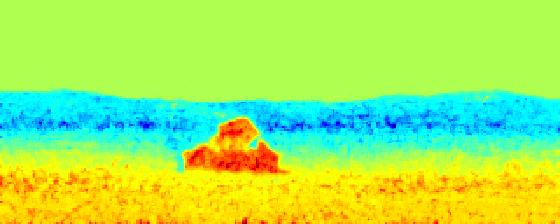}
	\includegraphics[width = .25\textwidth]{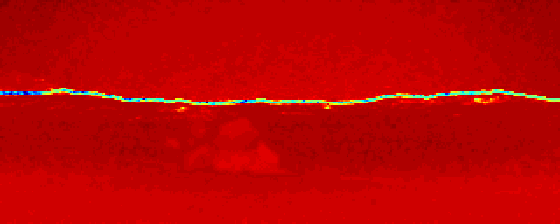}
	\includegraphics[width = .25\textwidth]{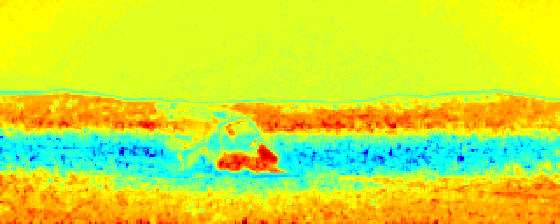}
	\vskip 0.005\textheight
	\includegraphics[width = .25\textwidth]{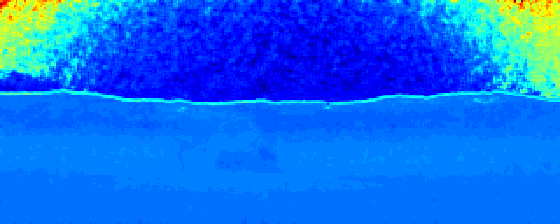}
	\includegraphics[width = .25\textwidth]{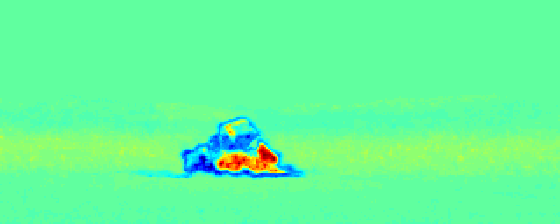}
	\includegraphics[width = .25\textwidth]{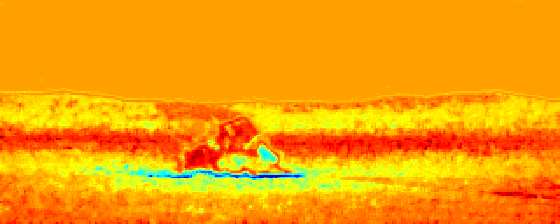}
	\vskip 0.005\textheight
	\includegraphics[width = .25\textwidth]{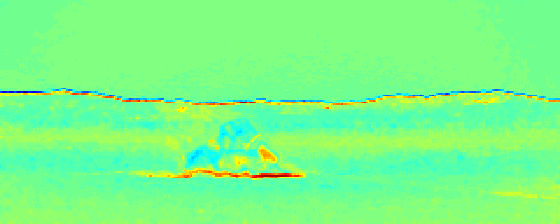}
	\includegraphics[width = .25\textwidth]{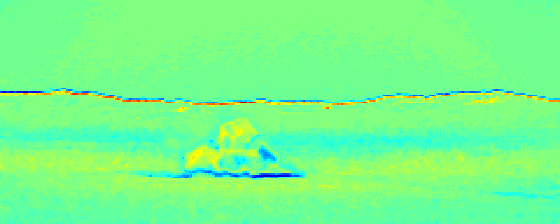}
	\includegraphics[width = .25\textwidth]{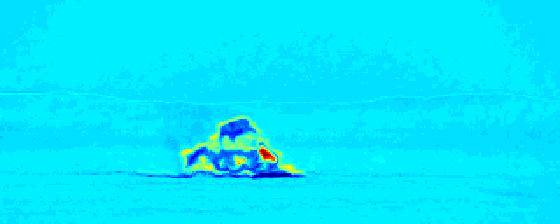}
	\vskip 0.005\textheight
	\includegraphics[width = .25\textwidth]{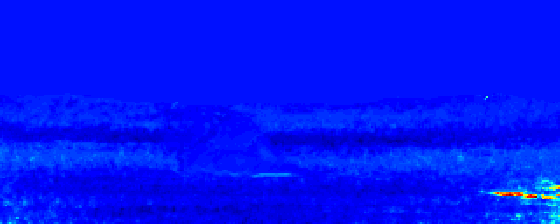}
	\includegraphics[width = .25\textwidth]{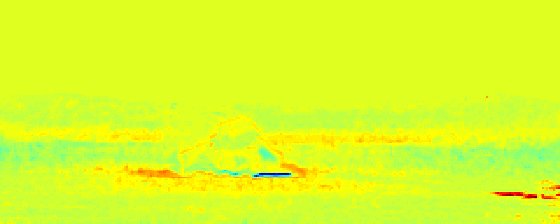}
	\includegraphics[width = .25\textwidth]{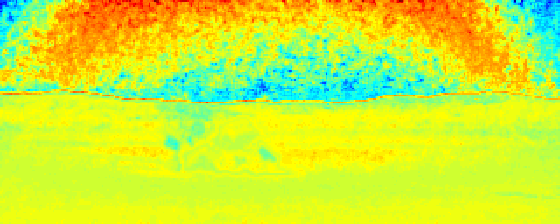}
  \caption{Spectral clustering result utilizing feature vectors: First 15 eigenvectors utilizing modified cosine distance metric.}
  \label{fig:SCCosFV}
\end{figure}

The next method is spectral clustering.  In the following figures we show the spectral clustering results.  The results consist of the first twenty eigenvectors of the normalized Laplacian matrix, shown in Figures \ref{fig:SCCos}--\ref{fig:SCCosFV}. Looking at the results, it would appear that the cosine distance metric and the Euclidean distance metric give quite similar clusters.  The clusters resulting from feature vectors are quite different, the borders between separate clusters are much more well defined.  It has also managed to isolate interesting areas of the plume itself, possibly being able to differentiate areas that are dust versus gas. It is interesting that all three of the results are able to differentiate different areas of the gas plume.

\subsection{Modified MBO}

\begin{figure}
\centering
    \includegraphics[width = .23\textwidth]{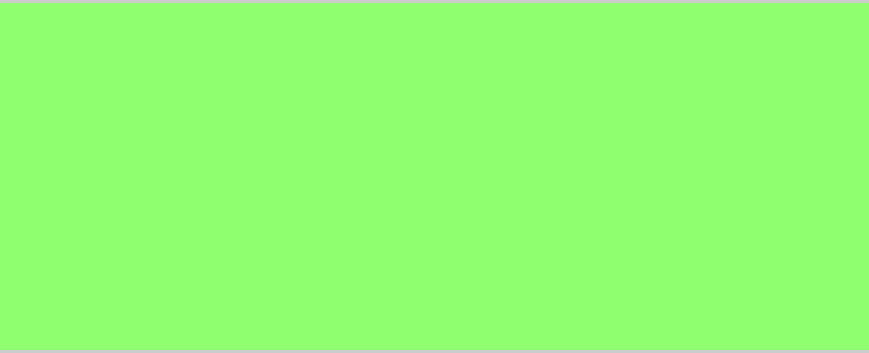}
    \includegraphics[width = .23\textwidth]{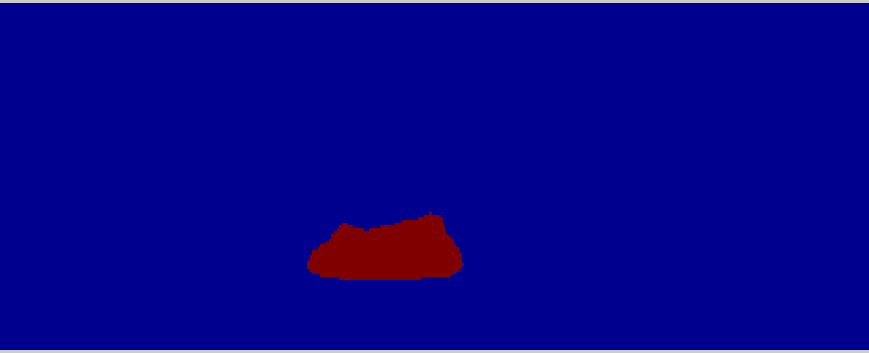}
    \includegraphics[width = .23\textwidth]{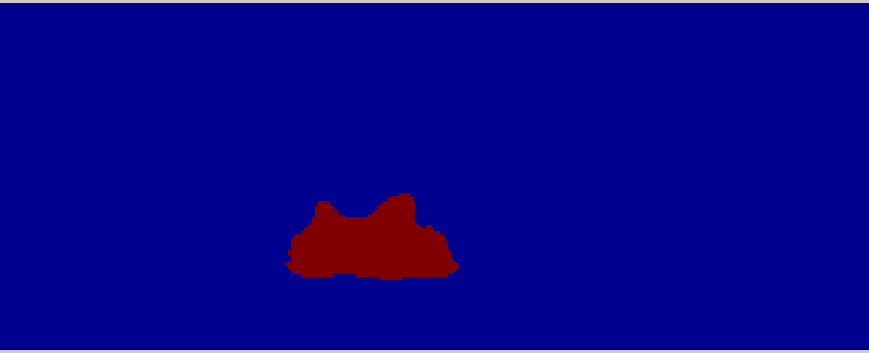}
    \includegraphics[width = .23\textwidth]{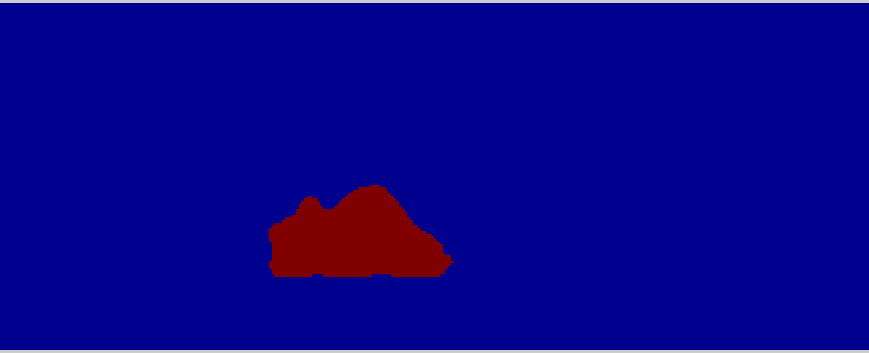}
    \vskip 0.005\textheight
    \includegraphics[width = .23\textwidth]{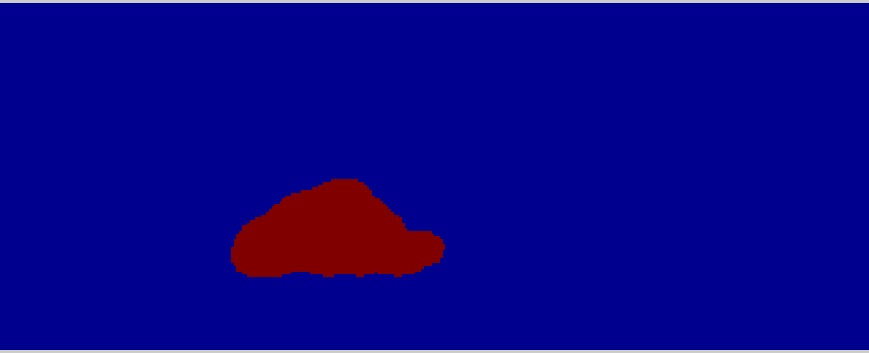}
    \includegraphics[width = .23\textwidth]{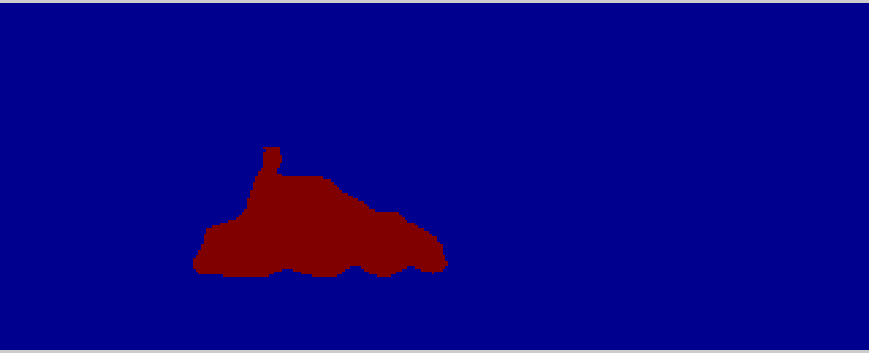}
    \includegraphics[width = .23\textwidth]{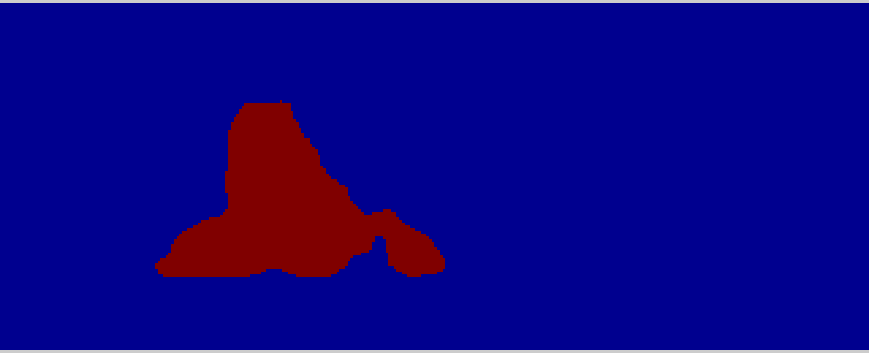}
    \includegraphics[width = .23\textwidth]{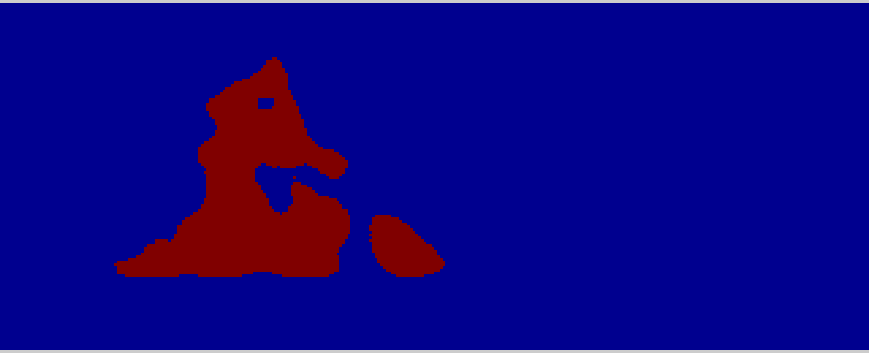}
    \vskip 0.005\textheight
    \includegraphics[width = .23\textwidth]{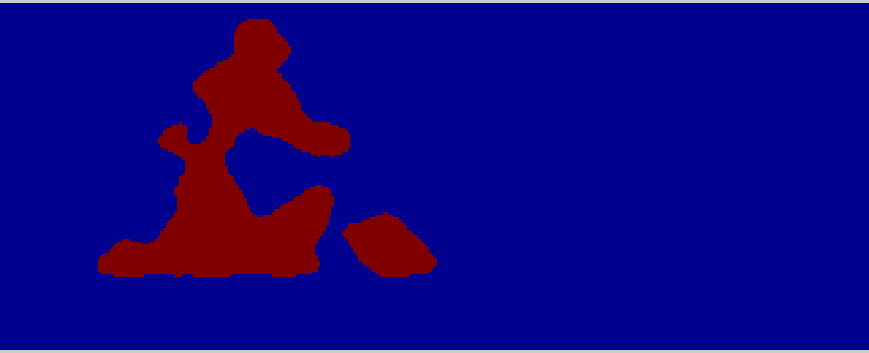}
    \includegraphics[width = .23\textwidth]{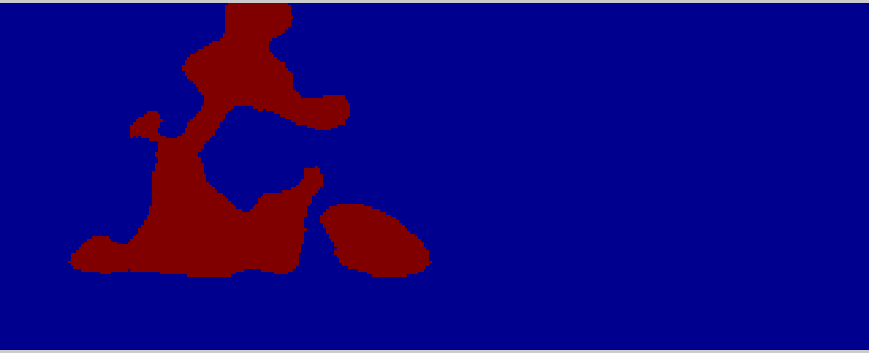}
    \includegraphics[width = .23\textwidth]{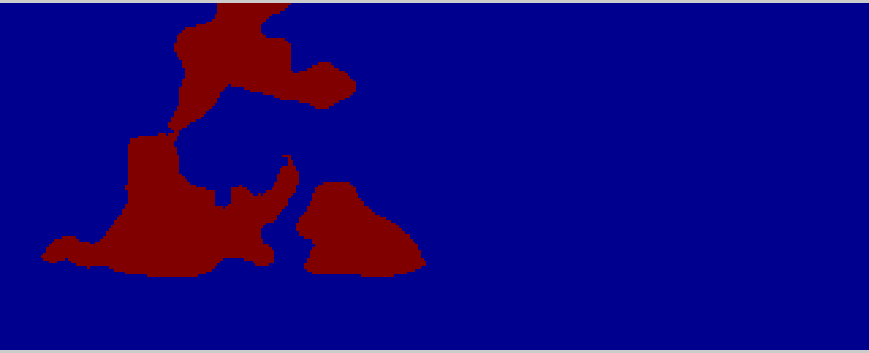}
    \includegraphics[width = .23\textwidth]{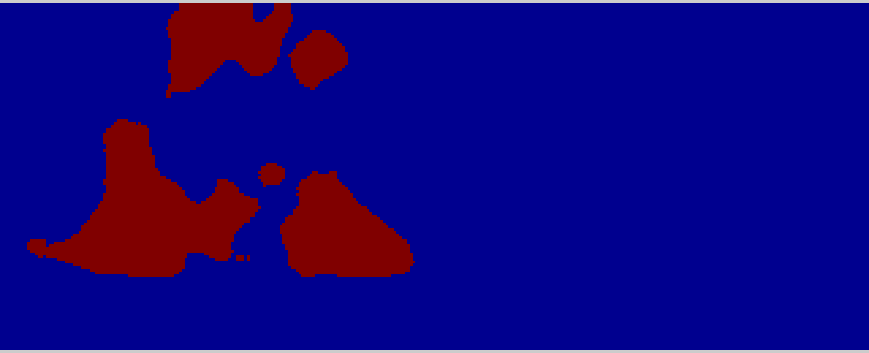}
    \vskip 0.005\textheight
    \includegraphics[width = .23\textwidth]{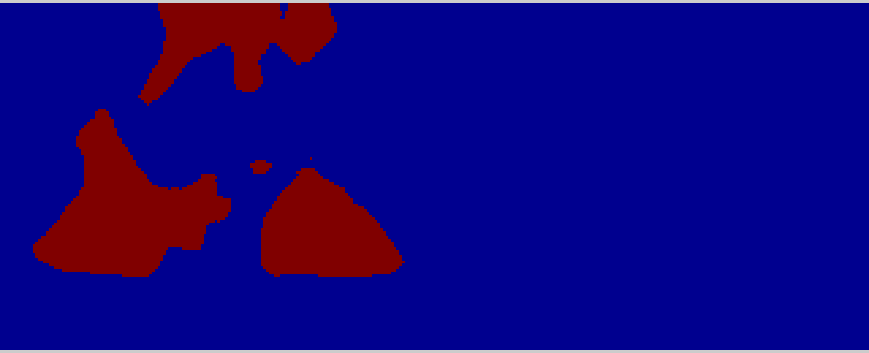}
    \includegraphics[width = .23\textwidth]{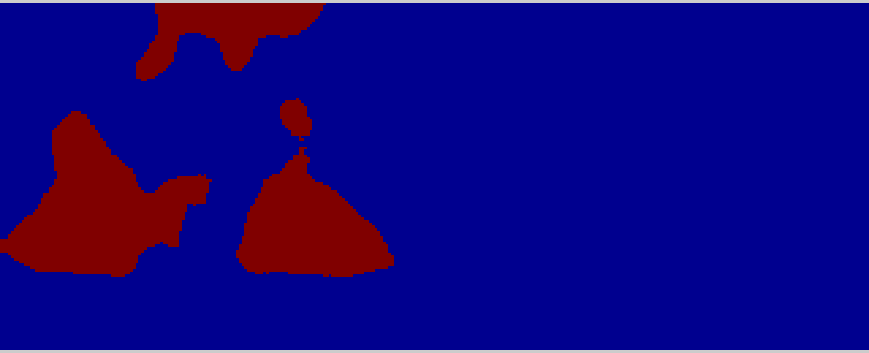}
    \includegraphics[width = .23\textwidth]{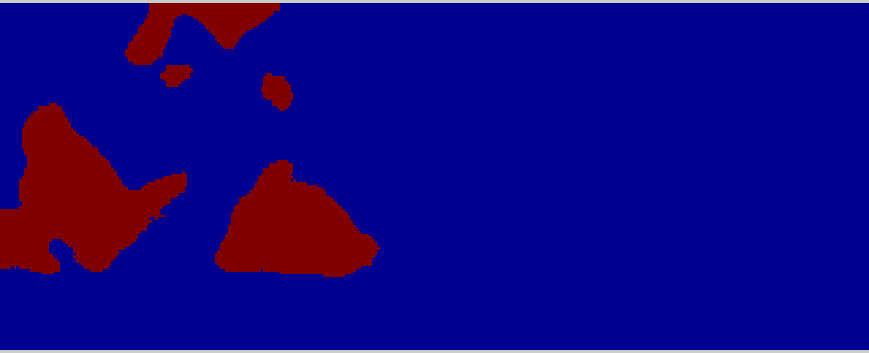}
    \includegraphics[width = .23\textwidth]{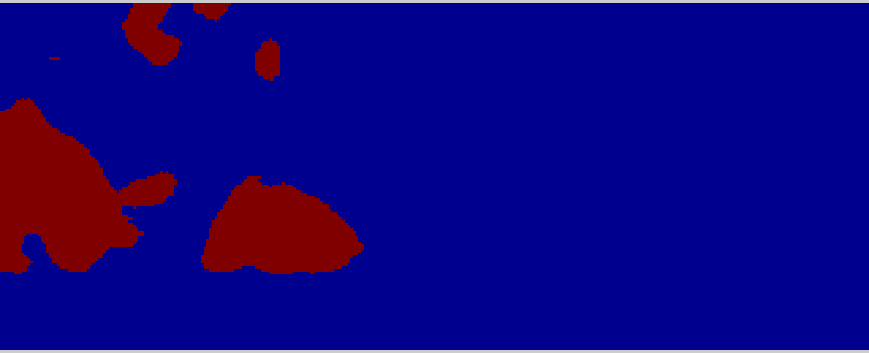}
    \vskip 0.005\textheight
    \includegraphics[width = .23\textwidth]{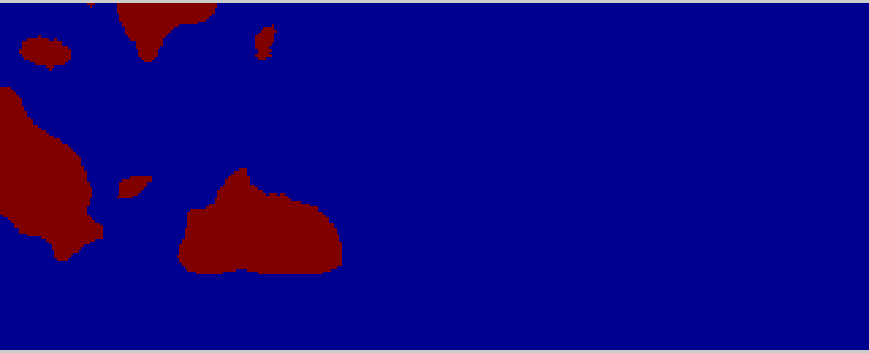}
    \includegraphics[width = .23\textwidth]{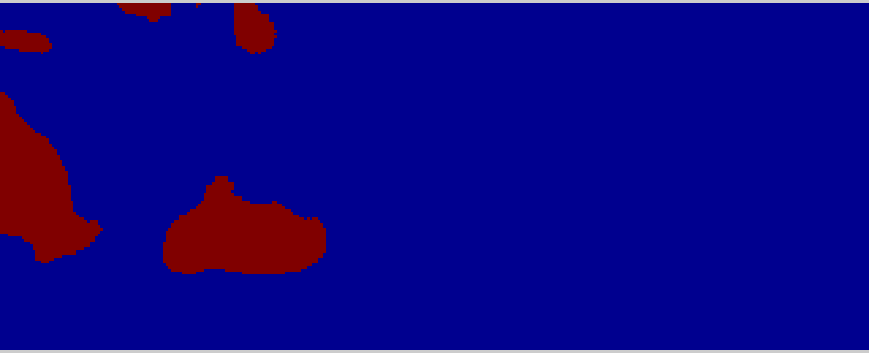}
    \includegraphics[width = .23\textwidth]{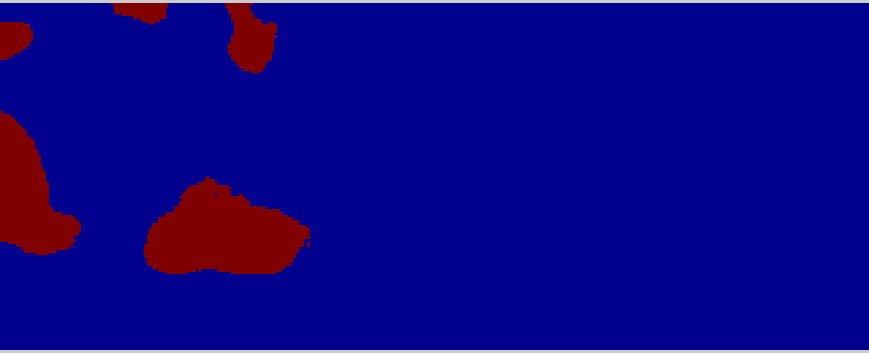}
    \includegraphics[width = .23\textwidth]{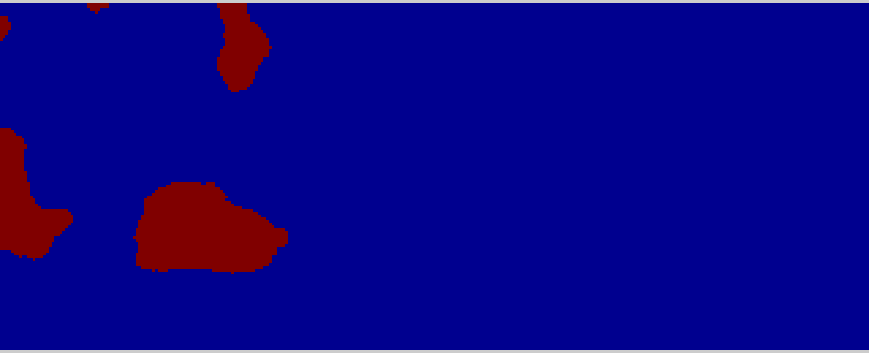}
  \caption{The initialization for the MBO scheme.}
  \label{fig:INI}
\end{figure}

\begin{figure}
  \centering
	\includegraphics[width = .23\textwidth]{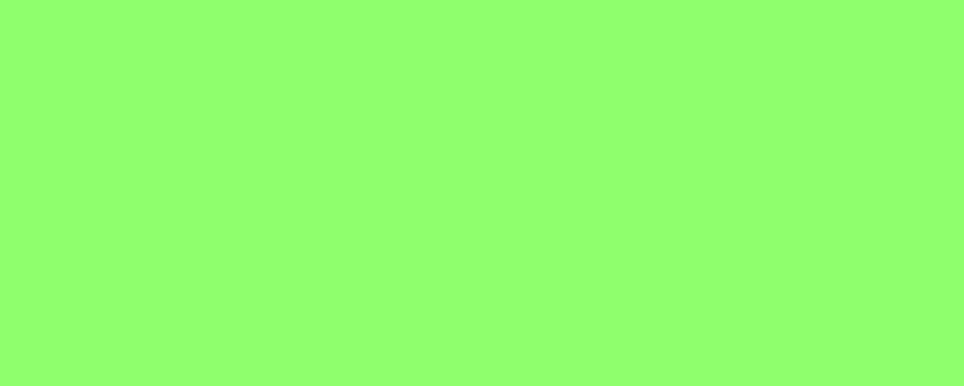}
	\includegraphics[width = .23\textwidth]{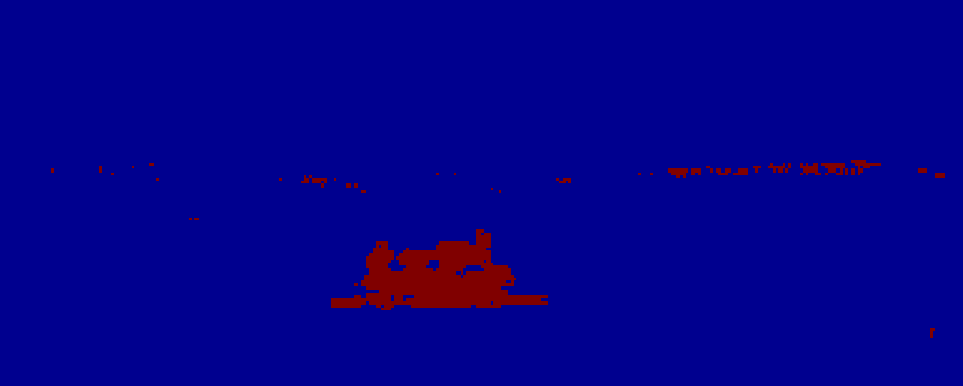}
	\includegraphics[width = .23\textwidth]{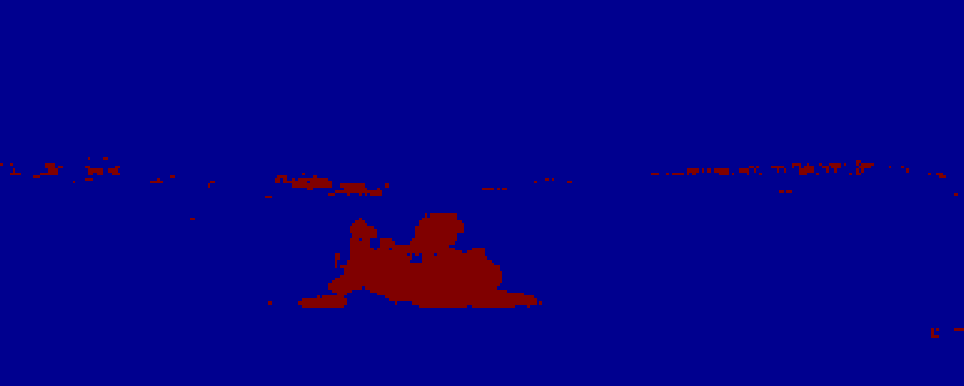}
	\includegraphics[width = .23\textwidth]{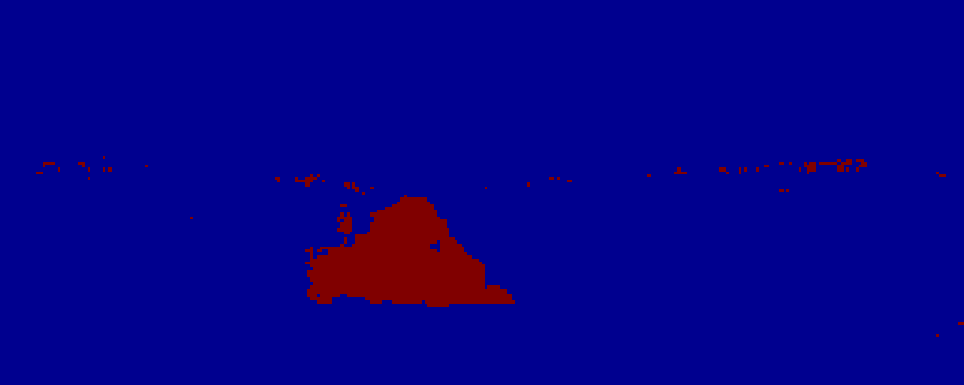}
	\vskip 0.005\textheight
	\includegraphics[width = .23\textwidth]{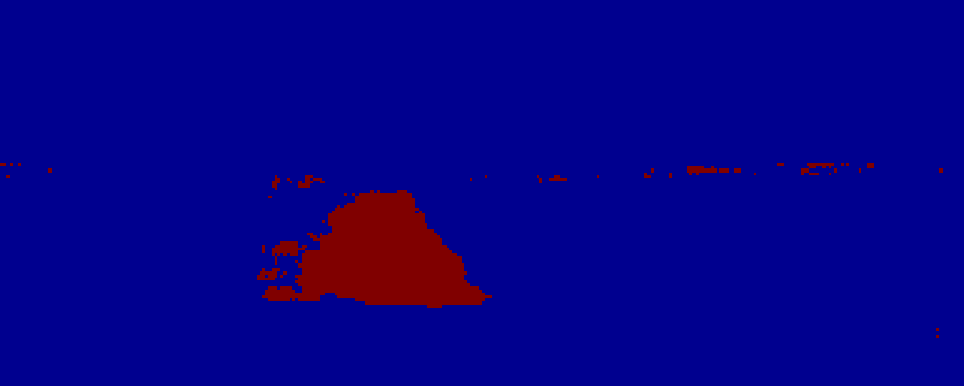}
	\includegraphics[width = .23\textwidth]{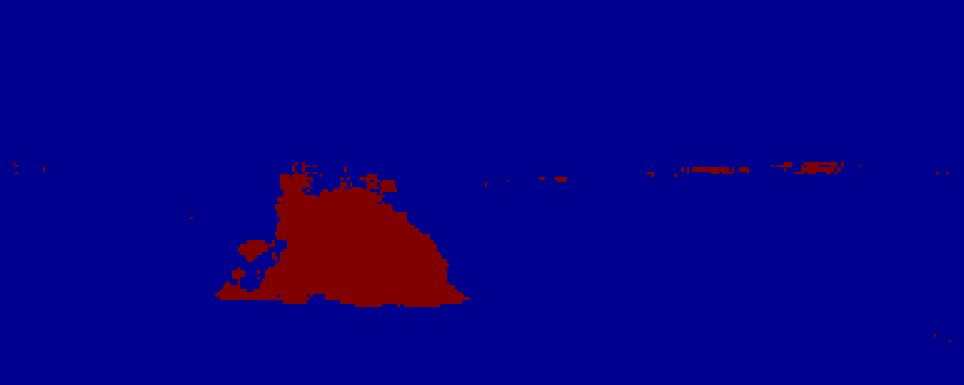}
	\includegraphics[width = .23\textwidth]{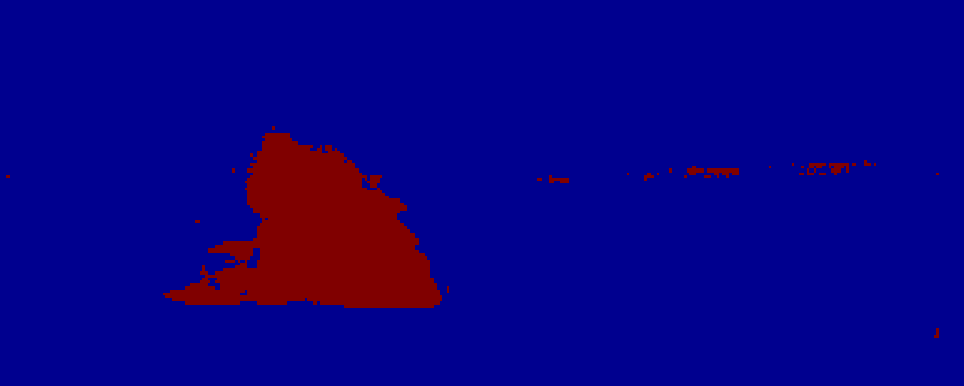}
	\includegraphics[width = .23\textwidth]{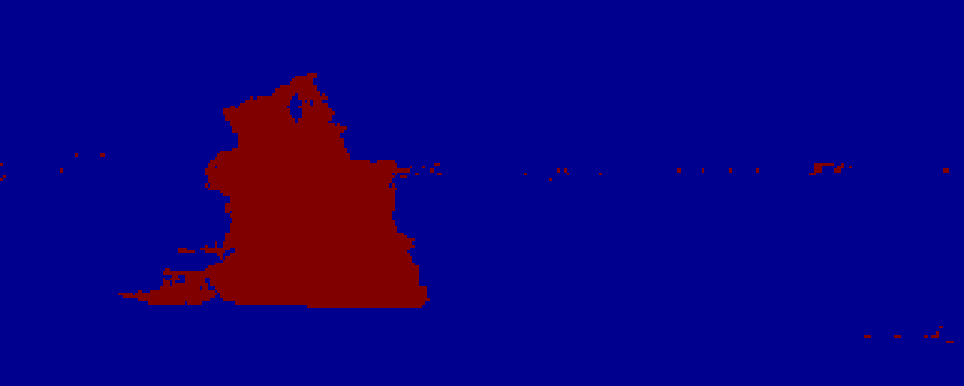}
	\vskip 0.005\textheight
	\includegraphics[width = .23\textwidth]{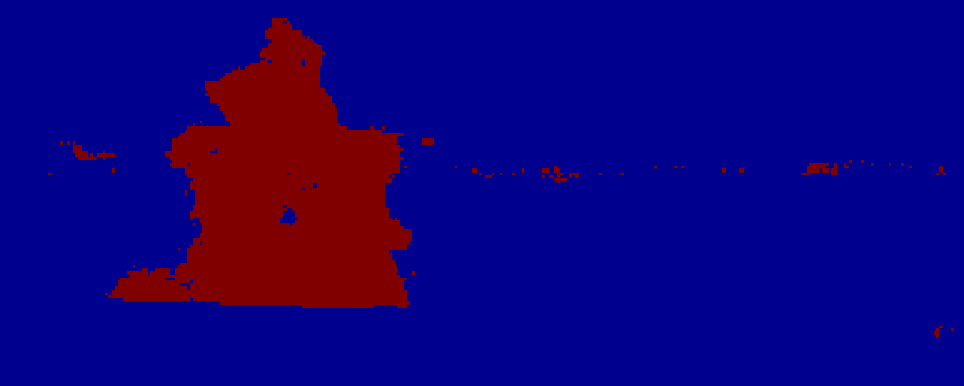}
	\includegraphics[width = .23\textwidth]{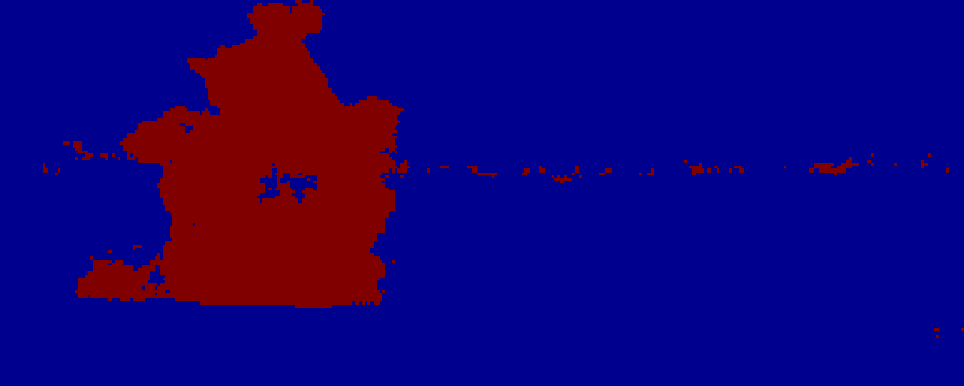}
	\includegraphics[width = .23\textwidth]{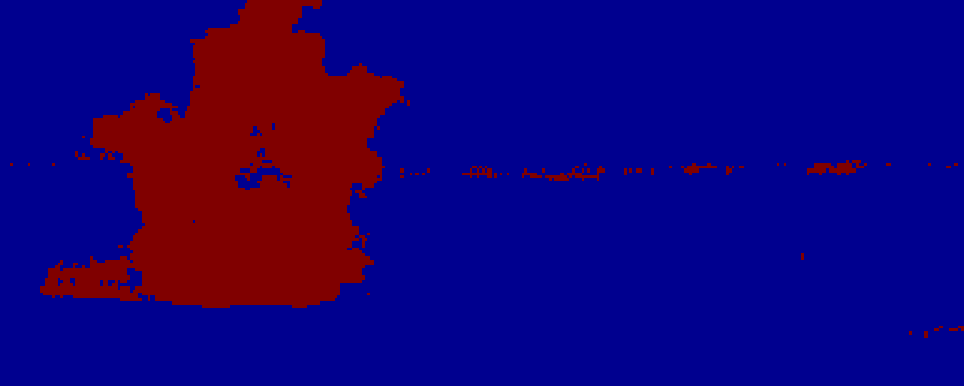}
	\includegraphics[width = .23\textwidth]{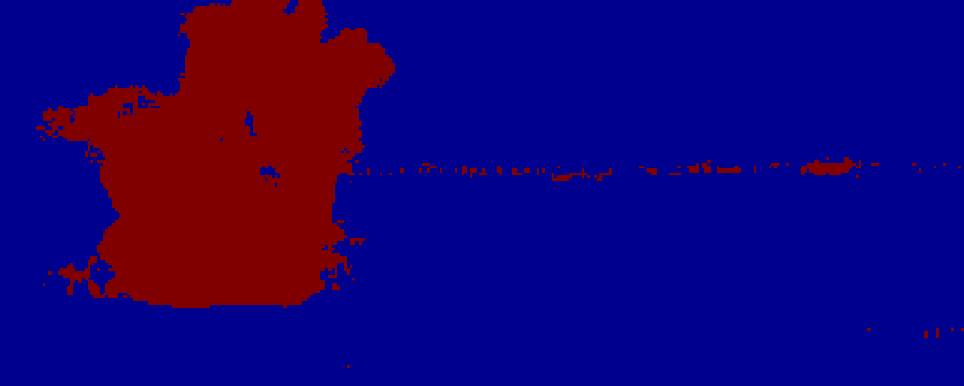}
	\vskip 0.005\textheight
	\includegraphics[width = .23\textwidth]{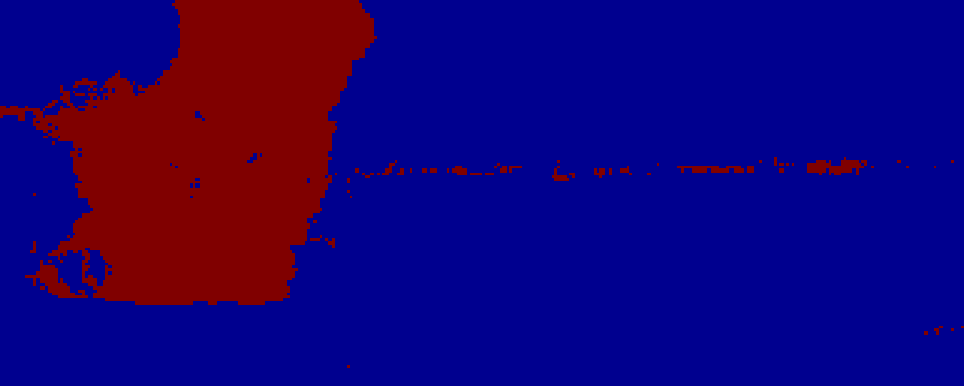}
	\includegraphics[width = .23\textwidth]{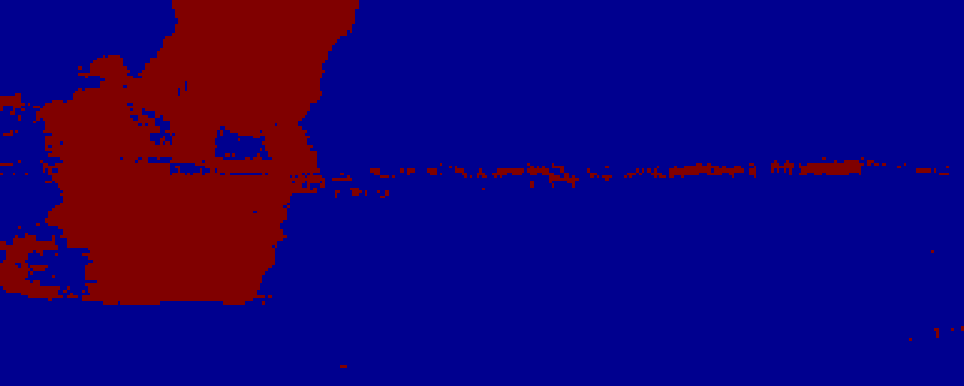}
	\includegraphics[width = .23\textwidth]{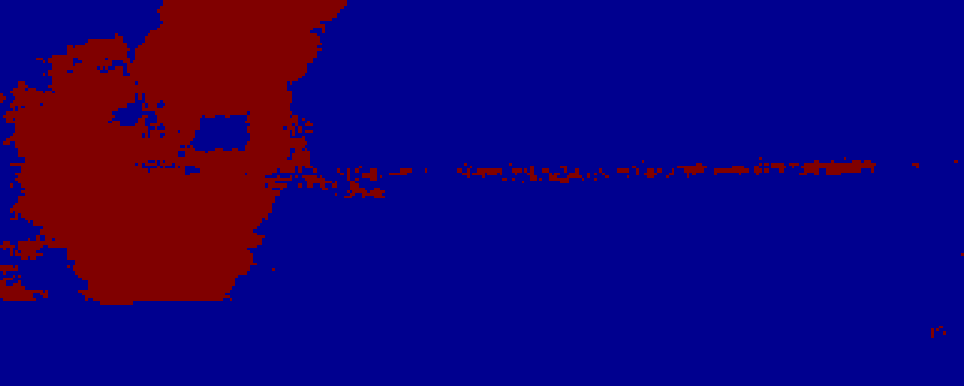}
	\includegraphics[width = .23\textwidth]{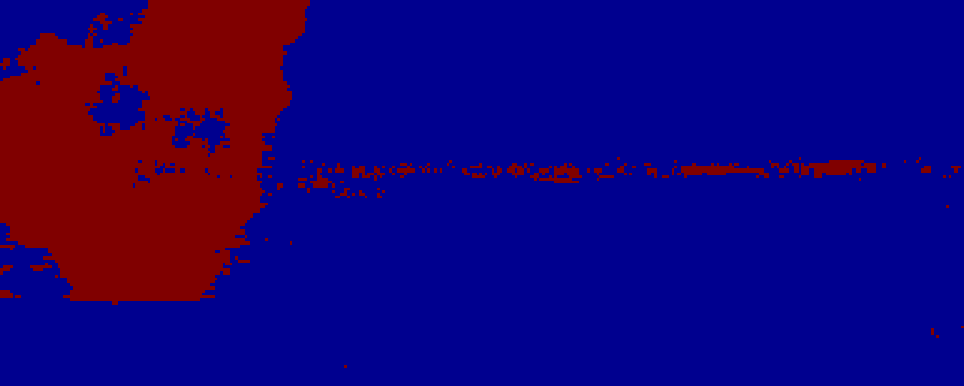}
	\vskip 0.005\textheight
	\includegraphics[width = .23\textwidth]{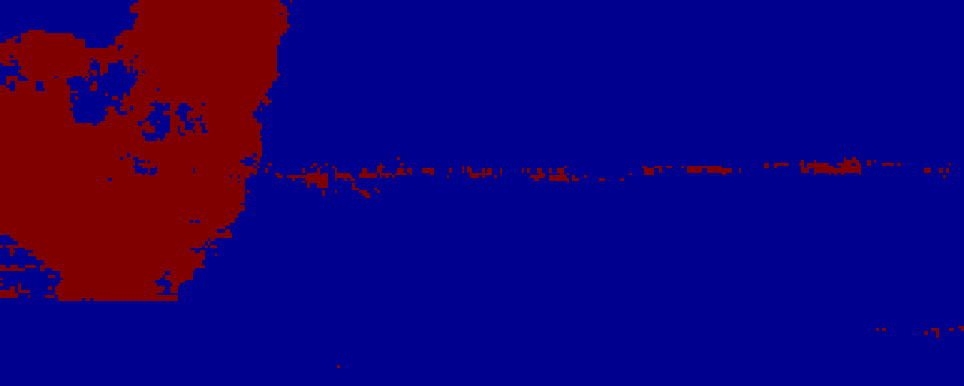}
	\includegraphics[width = .23\textwidth]{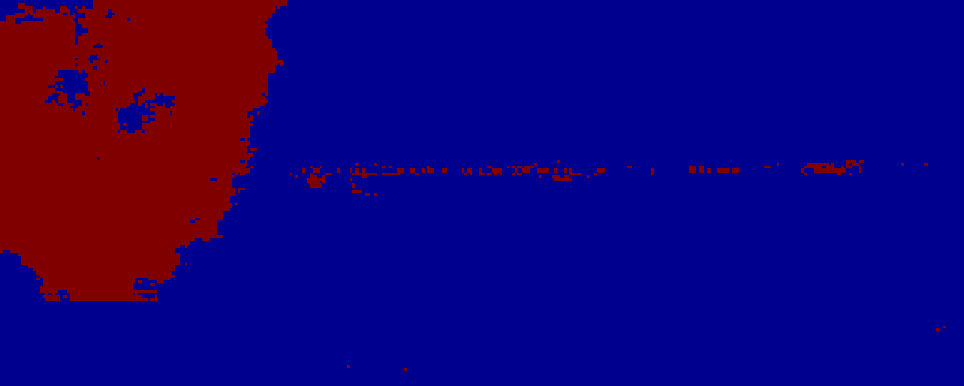}
	\includegraphics[width = .23\textwidth]{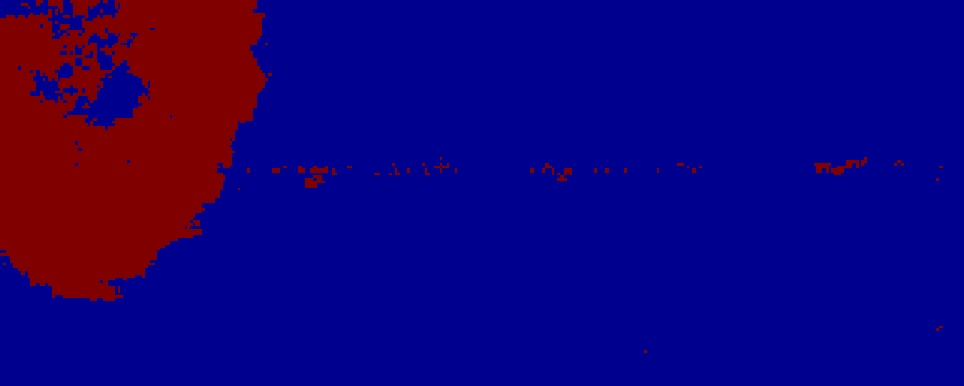}
	\includegraphics[width = .23\textwidth]{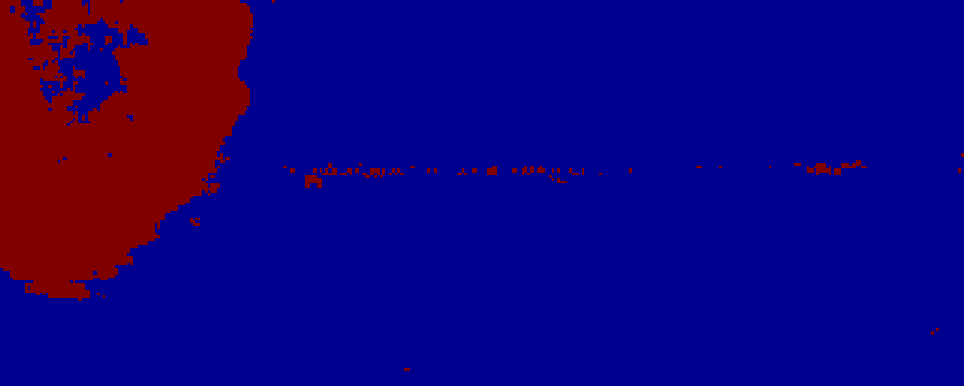}
  \caption{MBO result utilizing the first 5 spectral principal components.  The images show segmentation results starting one frame before the gas plume is released.}
  \label{fig:MBO5}
\end{figure}

\begin{figure}
  \centering
  	\includegraphics[width = .23\textwidth]{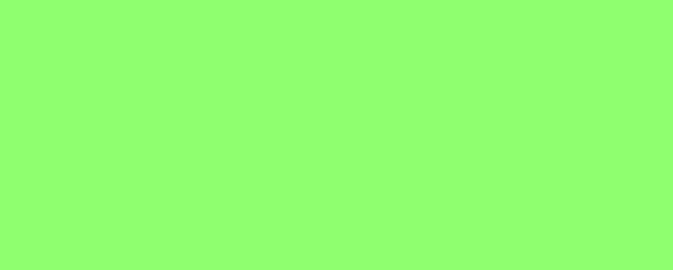}
	\includegraphics[width = .23\textwidth]{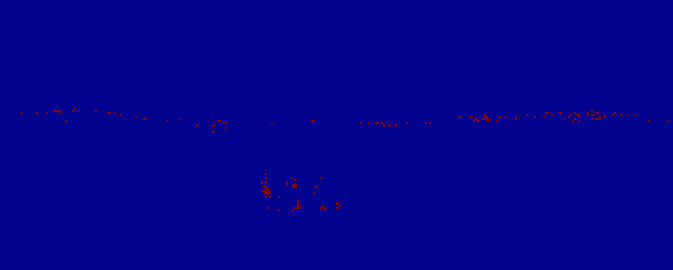}
	\includegraphics[width = .23\textwidth]{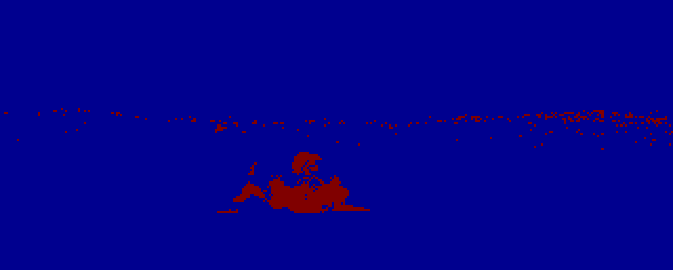}
	\includegraphics[width = .23\textwidth]{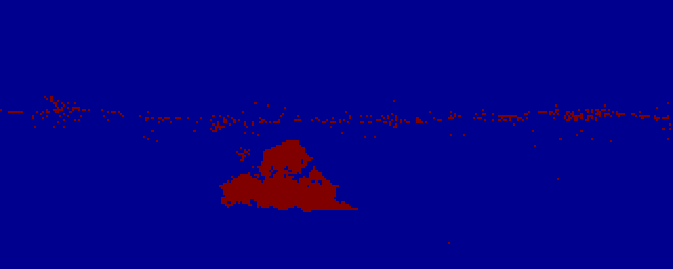}
	\vskip 0.005\textheight
	\includegraphics[width = .23\textwidth]{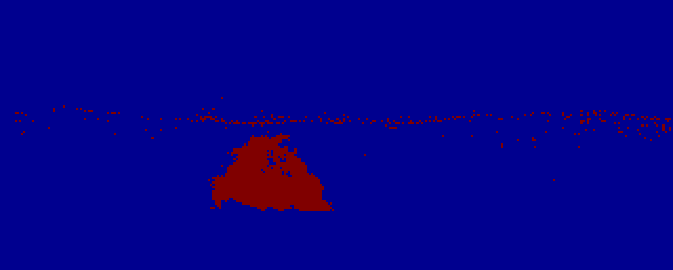}
	\includegraphics[width = .23\textwidth]{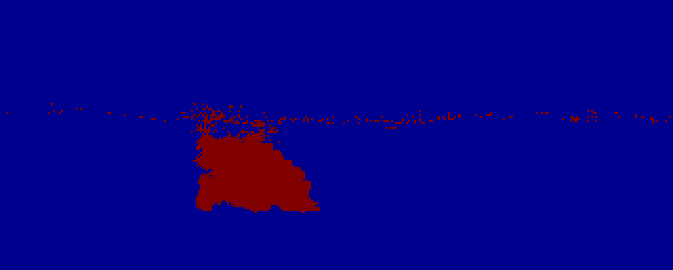}
	\includegraphics[width = .23\textwidth]{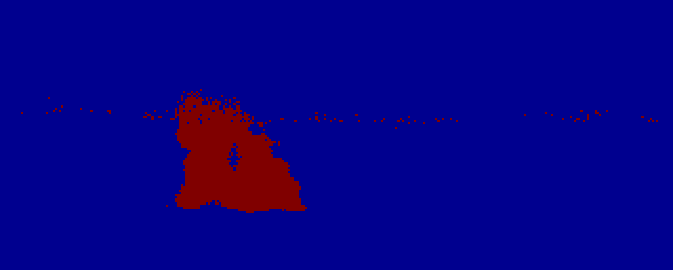}
	\includegraphics[width = .23\textwidth]{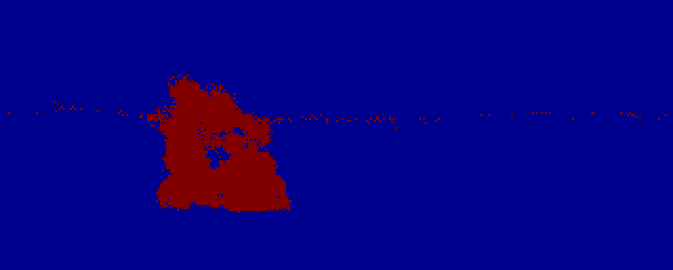}
	\vskip 0.005\textheight
	\includegraphics[width = .23\textwidth]{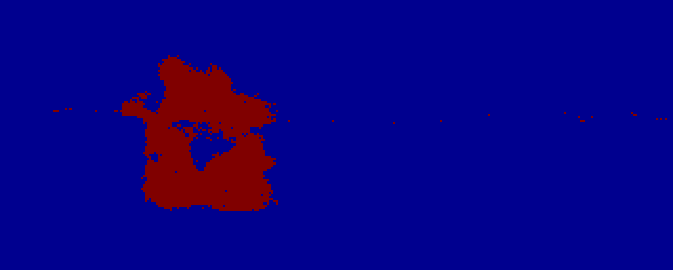}
	\includegraphics[width = .23\textwidth]{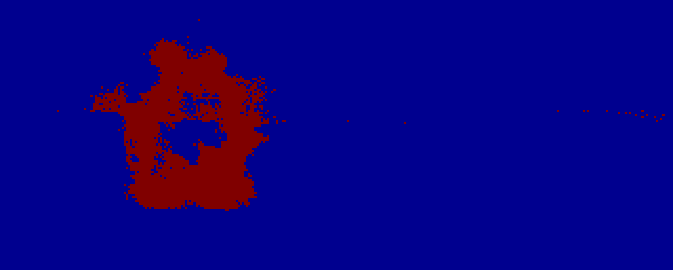}
	\includegraphics[width = .23\textwidth]{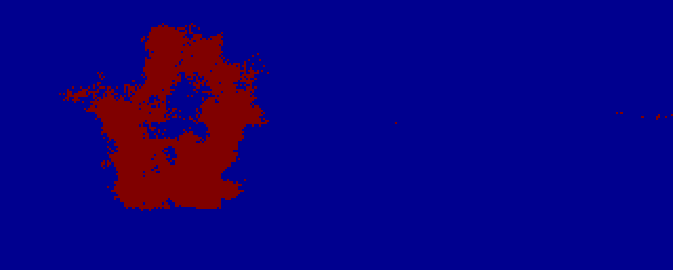}
	\includegraphics[width = .23\textwidth]{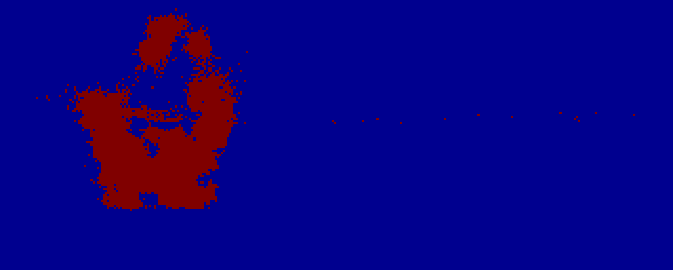}
	\vskip 0.005\textheight
	\includegraphics[width = .23\textwidth]{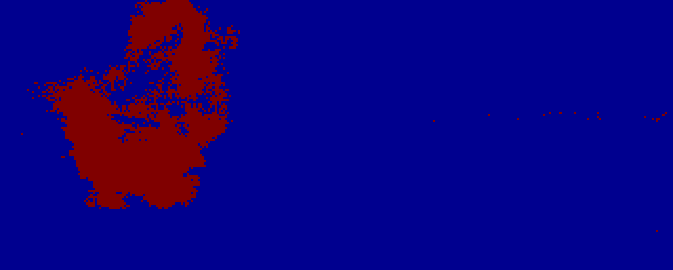}
	\includegraphics[width = .23\textwidth]{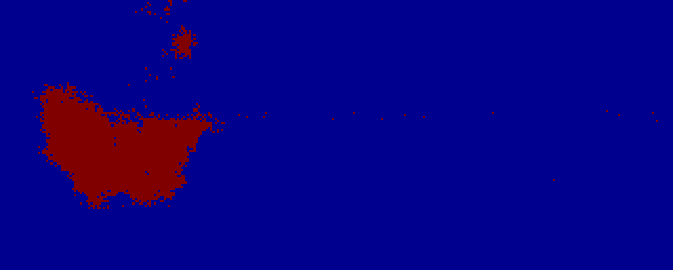}
	\includegraphics[width = .23\textwidth]{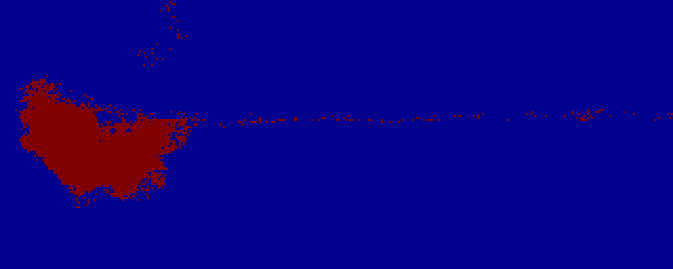}
	\includegraphics[width = .23\textwidth]{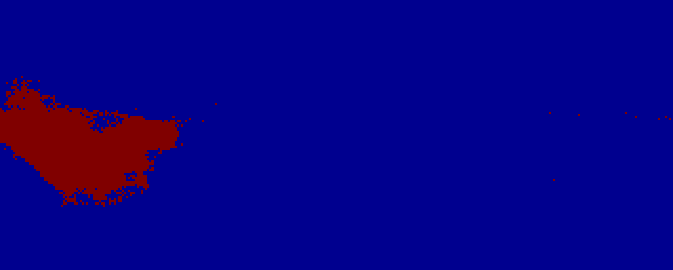}
	\vskip 0.005\textheight
	\includegraphics[width = .23\textwidth]{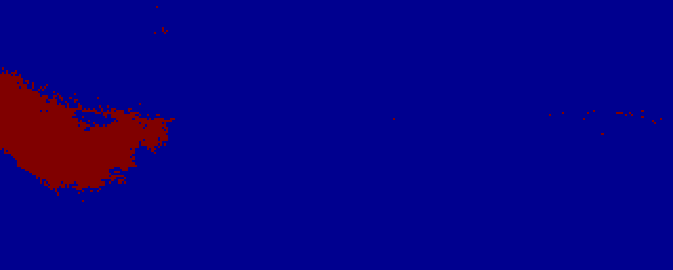}
	\includegraphics[width = .23\textwidth]{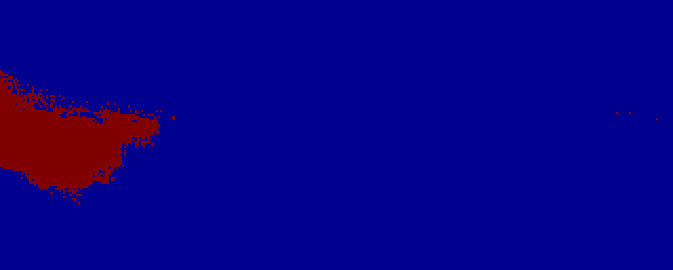}
	\includegraphics[width = .23\textwidth]{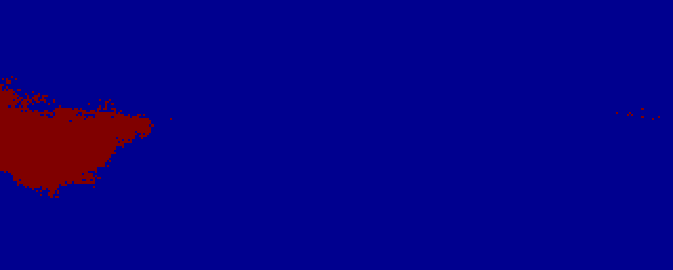}
	\includegraphics[width = .23\textwidth]{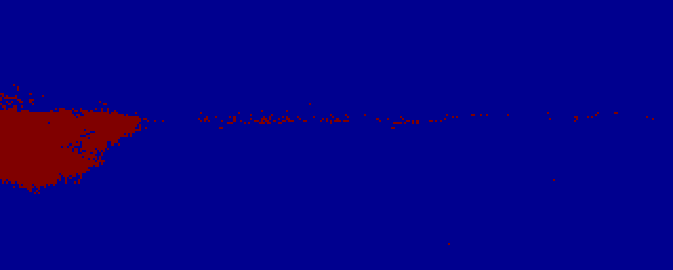}
  \caption{MBO result utilizing the first 1-3-5 spectral principal components.  The images show segmentation results starting one frame before the gas plume is released.}
  \label{fig:MBO3}
\end{figure}

This method captures subtle details in the movement of the plume between frames. The initialization was obtained by a basic background subtraction from the previous frame, and further cleaned by a 9 $\times$ 9 median filter. This left only the most concentrated sections of the plume, as seen in Figures \ref{fig:INI}. Now the 1-3-5 false color midway equalized video sequence as well as the video sequence obtained with the first five principal components and midway equalization are tested with the MBO scheme with results shown in Figure \ref{fig:MBO3} and Figure \ref{fig:MBO5}, respectively.  Using this initialization, the MBO scheme was able to differentiate the less concentrated areas of the chemical from the background. With the first five principal components, this algorithms is able to perform very well even when the plume becomes quite diffuse.  However, for the 1-3-5 false color video sequence, the diffusion in the later frames are unable to be segmented.  The graph Laplacian was constructed with each node as a pixel and the data used only spectral information rather than feature vectors.  Future work might use feature vectors for more spatial coherence.

\subsection{Remarks}
It is worth mentioning that both spectral clustering and the modified MBO scheme rely on the computation of eigenvectors of the graph Laplacian. Fast eigenvector solvers are essential to making these clustering algorithms computationally feasible. These solvers include the Nystr\"{o}m extension for approximating eigenvectors and the Rayleigh-Chebyshev\cite{RCAnderson} procedure for finding eigenvectors of large sparse Hermitian matrices. In our experiments, we used the Nystr\"{o}m extension; a detailed discussion and derivation of this method may be found in \cite{Nys2004}.

\section{ Conclusion and Future Work } \label{sec:CON}
In this work we demonstrated a systematic way of producing accurate and reliable false color RGB movies from LWIR hyperspectral video sequences. This was done though reliable dimension reduction techniques (such as PCA), and the midway method of histogram equalization. More modern methods of end member estimation such as non-negative matrix factorization (NNMF) may perform better at estimating background signatures \cite{MM2010}. The dimensionally reduced videos produced by this method were able to capture enough information for the segmentation of a chemical plume release. The results of K-means show that the cosine distance metric provides a better partitioning of the data. Also, our results show the utilization of feature vectors in each algorithm provided cleaner distinctions between clusters which resulted in better a segmentation of each frame. Spectral clustering was useful in differentiating levels of detail in the image. The MBO scheme that minimized the Ginzburg-Landau functional was able to quickly segment the plume in each frame. These results demonstrate the ability of dimension reduction and midway equalization to produce smooth, consistent false-color RGB video sequences, as well as the effectiveness of each segmentation method.  Contemporary segmentation methods, which utilize the graph Laplacian, with feature vectors provide an efficient means of segmenting chemical plume releases.

\section{Acknowledgements}
This work was an extension of results from the Johns Hopkins Applied Physics Laboratory. They provided us with background reading material on the physical processes associated with the problem and some preliminary MATLAB code. In particular, we would like to thank Dr. Josh Broadwater and Dr. Alison Carr for their contributions to this work. We also thank Professor Jean-Michel Morel for suggesting the idea of midway equalization and Cristina Garcia for assistance with spectral clustering algorithms. This project was funded by  NSF grants DMS-0914856, DMS-1118971, and DMS-1045536, Office of Naval Research (ONR) grants N000141210040 and N000141210838, and UC Lab Fees Research grant 12-LR-236660.

\vspace{.42 cm}

\noindent \begin{minipage}{\textwidth}

\end{minipage}

\end{document}